\documentclass[10pt]{article}

\usepackage[preprint]{tmlr}

\usepackage{amsmath,amsfonts,bm}

\def\eqref#1{equation~\ref{#1}}

\def\1{\bm{1}}

\DeclareMathAlphabet{\mathsfit}{\encodingdefault}{\sfdefault}{m}{sl}
\SetMathAlphabet{\mathsfit}{bold}{\encodingdefault}{\sfdefault}{bx}{n}

{}

\usepackage{amsmath,amssymb}
\usepackage{graphicx}
\usepackage{float}
\usepackage{booktabs}
\usepackage{longtable}
\usepackage{array}
\usepackage{tikz-cd}
\usepackage{multicol}
\usepackage{hyperref}
\usepackage{url}

\makeatletter
\newcommand{\leftalignedcaptionstyle}{%
  \long\def\@makecaption##1##2{%
    \vskip\abovecaptionskip
    \noindent\parbox{\linewidth}{\raggedright\normalfont ##1: ##2}\par
    \vskip\belowcaptionskip}}
\makeatother

\title{Perplexity Can Miss SAE Feature Damage Under Quantization}

\newcommand{\pythiamodel}{Pythia\mbox{-}70M}
\newcommand{\pythiadeduped}{Pythia\mbox{-}70M\mbox{-}deduped}
\newcommand{\gemmamodel}{Gemma\mbox{-}2\mbox{-}2B}

\author{\name Evan Duan \email evanduan@umich.edu \\
      \addr University of Michigan}

\def\month{MM}  
\def\year{YYYY} 
\def\openreview{\url{https://openreview.net/forum?id=XXXX}} 

\begin{document}
\raggedbottom

\maketitle

\begin{abstract}
Quantization is a standard path to deploying large language models, and quantized models are typically judged acceptable when perplexity or downstream accuracy remains close to the full-precision original. But behavioral parity need not imply feature fidelity: the sparse-autoencoder (SAE) features used to interpret a full-precision model may change after weight rounding. We test this directly by using a frozen SAE as a fixed measurement basis, encoding full-precision and round-to-nearest (RTN) quantized activations on identical tokens, and measuring per-feature survival by Pearson correlation across bit-widths from INT8 to INT4 on Pythia-70M and Gemma-2-2B. Our central finding is that perplexity can miss feature damage: on Gemma-2-2B, INT7 improves perplexity while degrading 18.7\% of active SAE features, and under sliding-window evaluation INT6 also improves perplexity while only 51.3\% of active features survive. Feature survival is graded rather than cliff-like, with 62.4\% of active Pythia features and 51.3\% of active Gemma features surviving at INT6; most non-surviving features are blurred rather than fully damaged. Survival is also predictable from full-precision feature statistics alone, with cross-validated AUC 0.92--0.97 and peak activation as the strongest marginal predictor. Finally, RTN quantization and matched-perplexity magnitude pruning damage strongly overlapping feature sets, with Jaccard overlap 0.79--0.86 and damage-score Spearman correlation 0.98. These results show that behavioral metrics alone are insufficient evidence that full-precision interpretability findings transfer to quantized models, motivating feature-level audits of compression.

\end{abstract}

\section{Introduction}

Quantization is a standard path to deploying large language models at scale: rounding weights from 16-bit floating point to 8, 4, or fewer bits can reduce memory and latency while preserving task-level performance under standard evaluation metrics~\citep{xiao2023smoothquant,frantar2022gptq,li2024evaluating}. That accounting is almost always behavioral. A quantized model is typically judged acceptable if its perplexity or downstream benchmark accuracy remains close to that of the full-precision original. Whether the model still computes in the same way---whether the internal features identified by interpretability research in the full-precision model survive rounding---is rarely tested.

This question is increasingly consequential. Sparse autoencoders (SAEs) have become a standard tool for decomposing language-model activations into interpretable features~\citep{cunningham2023sparse}, and a growing body of work builds analyses, safety audits, and steering interventions on top of features extracted from full-precision models~\citep{chalnev2024improving,o2024steering,bayat2025steering}. If those models are then deployed in quantized form, the features used to reason about model behavior may no longer be the features the deployed model actually uses. The reliability of interpretability under compression is therefore a precondition for interpretability being useful in deployment, yet it remains largely uncharacterized for quantization.

Prior work has introduced SAE feature survival as an audit tool for model compression under pruning~\citep{borobia2026pruning}. We use this instrument to ask a different deployment question: whether behavioral parity under quantization implies feature fidelity. Unlike pruning, quantization preserves all weights but lowers their precision, making it possible for perplexity to remain stable while the feature geometry read out by a full-precision SAE changes. Across \pythiamodel{} and \gemmamodel{}, we sweep RTN from 8 to 4 bits, predict feature survival from full-precision statistics, compare against pruning, and relate feature fidelity to perplexity.

We make the following empirical contributions:

    \textbf{Perplexity can miss feature damage.} On \gemmamodel{}, INT7 improves perplexity while degrading 18.7\% of active SAE features. Under sliding-window evaluation, INT6 also improves perplexity while only 51.3\% of active features survive, showing that task-level metrics can underestimate representational change.

    \textbf{Quantization and pruning damage similar features.} At similar perplexity regimes, RTN quantization and magnitude pruning affect strongly overlapping feature sets, with Jaccard overlap 0.79--0.86 and damage-score Spearman correlation 0.98. This suggests a shared pattern of compression-induced feature vulnerability despite different compression mechanisms.

    \textbf{Feature survival is graded.} As bit-width decreases, SAE features degrade systematically rather than failing all at once. At INT6, survival falls to 62.4\% on \pythiamodel{} and 51.3\% on \gemmamodel{}, with most non-surviving features degraded rather than fully damaged.
    
    \textbf{Feature survival is predictable.} INT6 survival can be predicted from full-precision feature statistics with cross-validated AUC 0.92 on \pythiamodel{} and 0.97 on \gemmamodel{}. Peak activation is the strongest marginal predictor: high-peak features survive reliably, while weak-signal features are most vulnerable to rounding-induced perturbation.

Together, these results show that behavioral metrics alone are insufficient evidence that full-precision interpretability findings transfer to quantized deployments. 

\section{Related Work}

\subsection{Sparse autoencoders and monosemantic features}\label{subsec:sparse-autoencoders-monosemantic-features}

Sparse autoencoders (SAEs) decompose the dense activation vectors of a language model into a larger set of sparse, individually interpretable features, offering a practical response to the superposition hypothesis under which networks represent more features than they have dimensions~\citep{elhage2022toy}. \citet{bricken2023towards} showed that SAEs trained on language-model activations recover monosemantic features, and \citet{cunningham2023sparse} demonstrated that the recovered features are highly interpretable.

\citet{templeton2024scaling} scaled the approach to a production model, and the Gemma Scope project~\citep{lieberum2024gemma} released pretrained residual-stream SAEs across the layers of the Gemma-2 family, providing the standardized dictionaries we use for our Gemma experiments. A growing body of work builds analyses, audits, and steering interventions on top of SAE features~\citep{chalnev2024improving,o2024steering,bayat2025steering}, typically using features extracted from full-precision models.

The reliability of SAEs as a measurement instrument is itself an active question. \citet{paulo2025sparse} show that SAEs trained on identical data with different random seeds learn substantially different features, and \citet{chanin2025sparse} show that feature recovery is sensitive to the sparsity hyperparameter. Recent methods aim to stabilize SAE training: \citet{jedryszek2026stable} add weight-regularization penalties that increase cross-seed feature sharedness and steering reliability, while other work encourages convergence across parallel or sequential SAE training runs~\citep{marks2024enhancing,martin2025attribution}. This line of work concerns variability introduced on the SAE-training side of the pipeline; it is complementary to our question, which concerns variability introduced on the model-compression side. In our experiments the SAE is held fixed and only the model weights change, so SAE-training variability is not the source of the feature changes we measure.

\subsection{Compression of language models}
Quantization and pruning are two dominant families of post-training compression. 
Quantization reduces numerical precision: round-to-nearest is the simplest scheme, while GPTQ and AWQ reduce quantization error more carefully~\citep{frantar2022gptq,lin2024awq}; mixed-precision and low-bit formats are now common in deployment~\citep{dettmers2022gpt3,dettmers2023qlora}. 
Pruning removes weights instead: magnitude pruning removes the smallest-magnitude weights~\citep{han2015learning}, while SparseGPT and Wanda provide scalable one-shot pruning methods for large language models~\citep{frantar2023sparsegpt,sun2024simple}. 
Both compression families are typically evaluated by perplexity or downstream accuracy; we instead measure feature-level representational change under quantization and use pruning as a matched-perplexity baseline.

\subsection{Interpretability under compression}

The intersection of SAE-based interpretability and model compression remains sparsely explored. The closest prior work is \citet{borobia2026pruning}, who study how unstructured pruning reshapes SAE features across multiple model families, pruning methods, and sparsity levels. They find that rare, low-firing features survive pruning better than frequent ones, that Wanda better preserves feature structure than magnitude pruning, and that geometric feature survival does not necessarily predict causal importance under ablation.

We build on this audit perspective but apply it to a question pruning cannot directly answer: whether a quantized model that remains behaviorally close to its full-precision original also preserves the SAE features used to interpret it. This changes the setting along three axes. First, we study quantization rather than pruning: weight rounding preserves all parameters but reduces their precision, while pruning removes weights entirely. Second, we use a frozen full-precision SAE on identical dense and compressed activations, measuring per-feature correlation directly rather than retraining and matching separate SAE dictionaries. Third, we vary bit-width from INT8 to INT4, tracing feature survival as a function of precision rather than sparsity.

\section{Methodology}\label{sec:methodology}

\begin{figure}[H]
    \centering
    \IfFileExists{prism-uploads/qdm_pipeline.png}{%
        \includegraphics[width=0.95\linewidth]{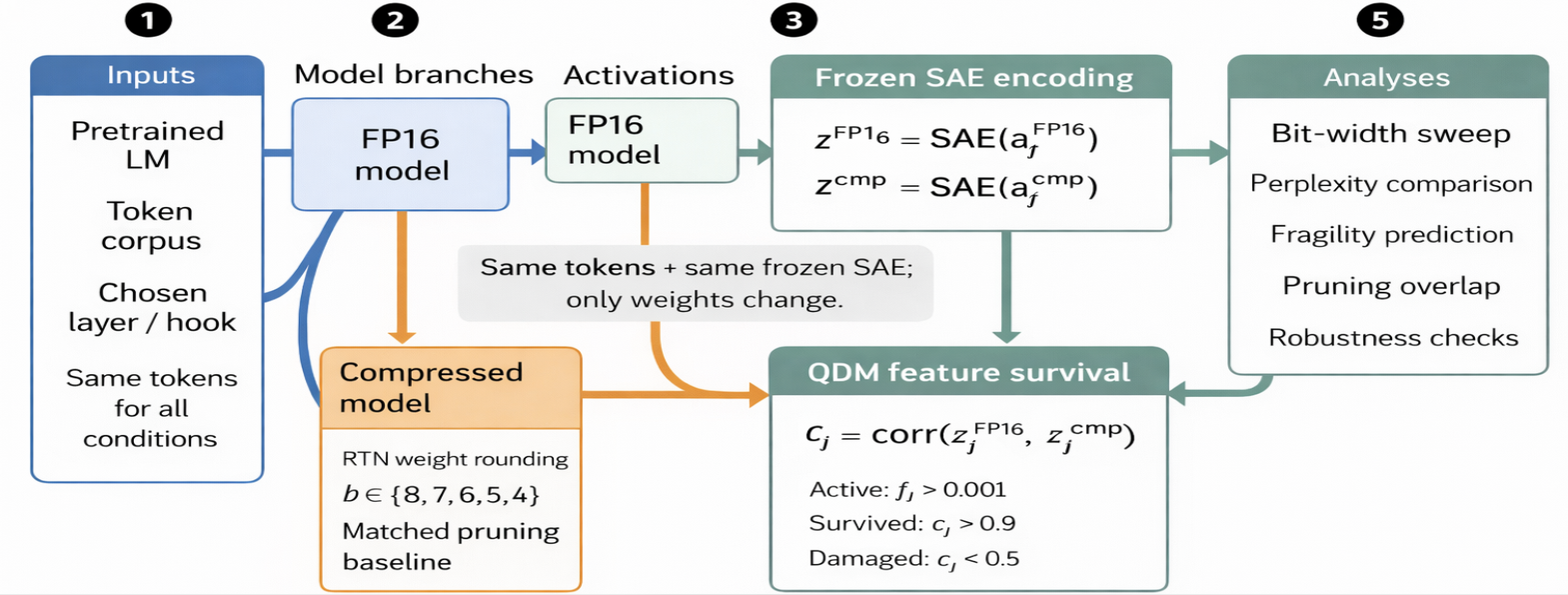}%
    }{%
        \fbox{\parbox{0.88\linewidth}{\centering QDM pipeline image file not found at \texttt{prism-uploads/qdm\_pipeline.png}.}}%
    }
    \caption{QDM pipeline: We encode FP16 and compressed activations with the same frozen SAE on identical tokens, measuring feature survival by per-feature correlation $c_j$.}\label{fig:qdm-pipeline-methodology}
\end{figure}

We measure how fixed SAE features change when the underlying language model is compressed. Across conditions, we keep the SAE, token set, and read-out site fixed, varying only the model weights. This section describes the models and SAEs, compression operators, feature-stability metric, behavioral evaluation, pruning baseline, fragility predictor, and stability checks. A compact glossary of notation and threshold conventions is provided in \ref{app:notation-glossary}

\subsection{Models, SAEs, and read-out site}\label{subsec:models-saes-readout}
We study two models spanning roughly a $30\times$ parameter range: \pythiadeduped{}~\citep{biderman2023pythia} and \gemmamodel{}~\citep{team2024gemma}. For each model, we use a publicly released residual-stream SAE at a fixed read-out layer: the \texttt{pythia-70m-deduped-res-sm} SAE at \texttt{blocks.4.hook\_resid\_post} for \pythiamodel{}, and the Gemma Scope layer-12 residual-stream SAE with width $d_{\mathrm{sae}}=16384$ for \gemmamodel{}. \ref{app:model-sae-evaluation-configuration} summarizes the model, SAE, token-budget, and evaluation configuration.

Let $h^{(l)}(t;\theta)$ denote the residual-stream activation of model $f_\theta$ at layer $l$ and token position $t$. A fixed SAE encoder $E$ maps this activation into feature space:
\begin{equation}
z(t;\theta)=E\left(h^{(l)}(t;\theta)\right),
\label{eq:sae-activation}
\end{equation}
where $z_j(t;\theta)$ is feature $j$'s activation. Across all compression conditions, $E$, the token set, and the read-out layer are fixed; only the model parameters $\theta$ change.

\subsection{Compression operators}\label{subsec:compression-operators}

We consider two families of weight compression. Both are applied to transformer-block linear weights: attention projections, MLP projections, and, for \gemmamodel{}'s gated MLP, the gate projection. Layer-norm and embedding parameters are left in full precision. Exact module-name patterns and exclusions are listed in \ref{app:quantized-module-implementation-details}.

\paragraph{Round-to-nearest quantization (RTN).}
For each targeted tensor $W$, we apply per-output-channel RTN quantization. For bit-width $b$, with signed range $q_{\min}=-2^{b-1}$ and $q_{\max}=2^{b-1}-1$, each output channel is scaled by its maximum absolute weight and quantized as
\begin{equation}
\widehat{W}_{i,c}
=
s_c\,\operatorname{clip}\left(
\operatorname{round}\left(\frac{W_{i,c}}{s_c}\right),
q_{\min},q_{\max}
\right),
\qquad
s_c=\frac{\max_i |W_{i,c}|}{q_{\max}}.
\label{eq:rtn-quantize}
\end{equation}
We sweep $b\in\{8,7,6,5,4\}$ and dequantize weights back into floating-point tensors, simulating low-bit rounding without changing the model architecture.

\paragraph{Magnitude pruning.}
We zero the smallest-magnitude fraction $p$ of weights in each targeted tensor and use pruning only as a matched-perplexity baseline, calibrated as described in Section~\ref{subsec:matched-perplexity-pruning-baseline}.

\subsection{Feature-stability metric}\label{sec:feature-stability}

We quantify feature stability by comparing each SAE feature's activation pattern before and after compression. 
For feature $j$, let 
$x_t = z_j(t;\theta_{\mathrm{FP16}})$ and 
$y_t = z_j(t;\theta_C)$ denote its full-precision and compressed activations at token position $t$ over a shared token set of size $N$. 
We define the feature-stability score as the Pearson correlation
\begin{equation}
c_j =
\frac{\sum_t (x_t-\bar{x})(y_t-\bar{y})}
{\sqrt{\sum_t {\left(x_t-\bar{x}\right)}^{2}
\sum_t {\left(y_t-\bar{y}\right)}^{2}}},
\label{eq:feature-correlation}
\end{equation}
where $\bar{x}$ and $\bar{y}$ are means over token positions. 
Since both activations are evaluated on identical token positions with the same frozen SAE, this comparison isolates the compression intervention from token-sampling variation. 
To avoid storing the full $N \times d_{\mathrm{sae}}$ activation matrix, we compute the same correlation from streaming sufficient statistics; implementation details are provided in \ref{app:streaming-estimator}.

We restrict survival statistics to features that are active in the full-precision model. 
Feature $j$ is active if its FP16 firing rate
\begin{equation}
f_j =
\frac{1}{N}\sum_t \mathbf{1}\!\left[z_j(t;\theta_{\mathrm{FP16}})>0\right]
\label{eq:firing-rate}
\end{equation}
exceeds $0.001$.

\subsection{Survival taxonomy}\label{sec:survival-taxonomy}

We summarize the distribution of $\{c_j\}$ over active features with three bands, using a survival threshold $t_s=0.9$ and a damage threshold $t_d=0.5$:
\begin{equation}
\begin{aligned}
\text{survived:} \quad & c_j > t_s 
&& \text{strongly aligned with the FP16 activation pattern},\\
\text{degraded:} \quad & t_d \leq c_j \leq t_s 
&& \text{partially aligned},\\
\text{damaged:} \quad & c_j < t_d 
&& \text{weakly aligned under the frozen SAE basis}.
\end{aligned}
\label{eq:survival-taxonomy}
\end{equation}
We use $t_s=0.9$ and $t_d=0.5$ as default thresholds and evaluate sensitivity to alternative cutoffs in \ref{app:stability-ablation-threshold-sensitivity}.

\subsection{Behavioral evaluation}\label{sec:behavioral-evaluation}

We report token-level perplexity on WikiText-2-raw using two protocols. 
In the main feature-extraction pipeline, we use a fixed chunked protocol: the token stream is split into non-overlapping blocks of length $L$ ($L=512$ for Pythia and $L=256$ for Gemma), and mean autoregressive cross-entropy is computed over scored positions within each block. Because each block resets context, this protocol can inflate absolute perplexity, but it is held fixed across compression conditions and therefore supports within-protocol perplexity deltas.

For Gemma, we additionally run a sliding-window behavioral check with window size $W=2048$ and stride $S=512$. This evaluation uses a HuggingFace-format Gemma model with the same per-output-channel RTN scheme and is used only to test whether perplexity trends are robust to a stronger behavioral evaluation protocol. The SAE feature analysis itself uses the TransformerLens/Gemma Scope activation-extraction pipeline. Full sliding-window details are provided in \ref{app:sliding-window}.

For any condition $C$, we report the relative perplexity change
\begin{equation}
\Delta_{\mathrm{PPL}}(C)=\frac{\mathrm{PPL}(C)}{\mathrm{PPL}(\mathrm{FP16})}-1.
\label{eq:ppl-delta}
\end{equation}

\subsection{Matched-perplexity pruning baseline}\label{subsec:matched-perplexity-pruning-baseline}

To compare quantization and pruning at a similar behavioral cost, we calibrate magnitude-pruning sparsity to the RTN INT6 perplexity regime. Let 
\(\mathrm{PPL}^{\star}=\mathrm{PPL}(\mathrm{RTN\ INT6})\). We choose the pruning sparsity
\begin{equation}
p^{\star}
=
\arg\min_p
\left|
\mathrm{PPL}(\operatorname{prune}(\theta,p))
-
\mathrm{PPL}^{\star}
\right|.
\label{eq:matched-pruning-search}
\end{equation}
The sparsity search and achieved perplexity matches are reported in \ref{app:pruning-calibration}. The calibrated pruning condition is then analyzed with the same frozen-SAE feature pipeline as RTN INT6.

We compare RTN INT6 and pruning using two per-feature overlap measures: the Jaccard overlap of non-survived feature sets,
\[
J=\frac{|A\cap B|}{|A\cup B|},
\]
where \(A\) and \(B\) are the features with \(c_j \leq t_s\) under each method, and the Spearman correlation of damage scores \(d_j=1-c_j\) across active features.

\subsection{Fragility predictor}\label{subsec:fragility-predictor}

To test whether survival is predictable from full-precision statistics alone, we fit an L2-regularized logistic regression predicting the INT6 survival outcome 
$y_j=\mathbf{1}[c_j>t_s]$. 
The predictors are four FP16 feature statistics: rarity 
$-\log(f_j+\epsilon)$, log mean activation 
$\log(\mu_j+\epsilon)$, log peak activation 
$\log(\max_t z_j(t;\theta_{\mathrm{FP16}})+\epsilon)$, and log concentration
\[
\log\left(
\frac{\max_t z_j(t;\theta_{\mathrm{FP16}})+\epsilon}{\mu_j+\epsilon}
\right),
\]
where $f_j$ is the firing rate, 
$\mu_j=(1/N)\sum_t z_j(t;\theta_{\mathrm{FP16}})$, and $\epsilon$ is a small numerical-stability constant. 
After standardizing predictors, we fit $P(y_j=1)=\sigma(\beta_0+\beta^\top\phi_j)$ with inverse regularization strength $C=1$. We report 5-fold stratified cross-validated AUC and standardized coefficients, with additional diagnostics in \ref{app:fragility-predictor-details}. Because the predictors are correlated, we interpret coefficients jointly rather than as isolated causal effects.

\subsection{Stability protocol}\label{subsec:stability-protocol}

Unless stated otherwise, we use 200k tokens for \pythiamodel{} and 500k tokens for \gemmamodel{}. On Pythia INT6, we test token-budget sensitivity, random-subset stability, and an FP16-vs-FP16 null; we also repeat the sweep at a second layer to check layer sensitivity.

\section{Results}\label{sec:results}

We report results for \pythiamodel{} and \gemmamodel{} using the fixed-SAE protocol from Section~\ref{sec:methodology}. Survival, degradation, and damage are computed over FP16-active features using the thresholds in Section~\ref{sec:survival-taxonomy}; all correlations use identical token positions, isolating the effect of weight compression.

\subsection{Feature survival under quantization is graded and consistent across scale}\label{subsec:feature-survival-quantization}

\begin{figure}[t]
    \centering
    \begin{minipage}{0.49\linewidth}
        \centering
        \includegraphics[width=\linewidth]{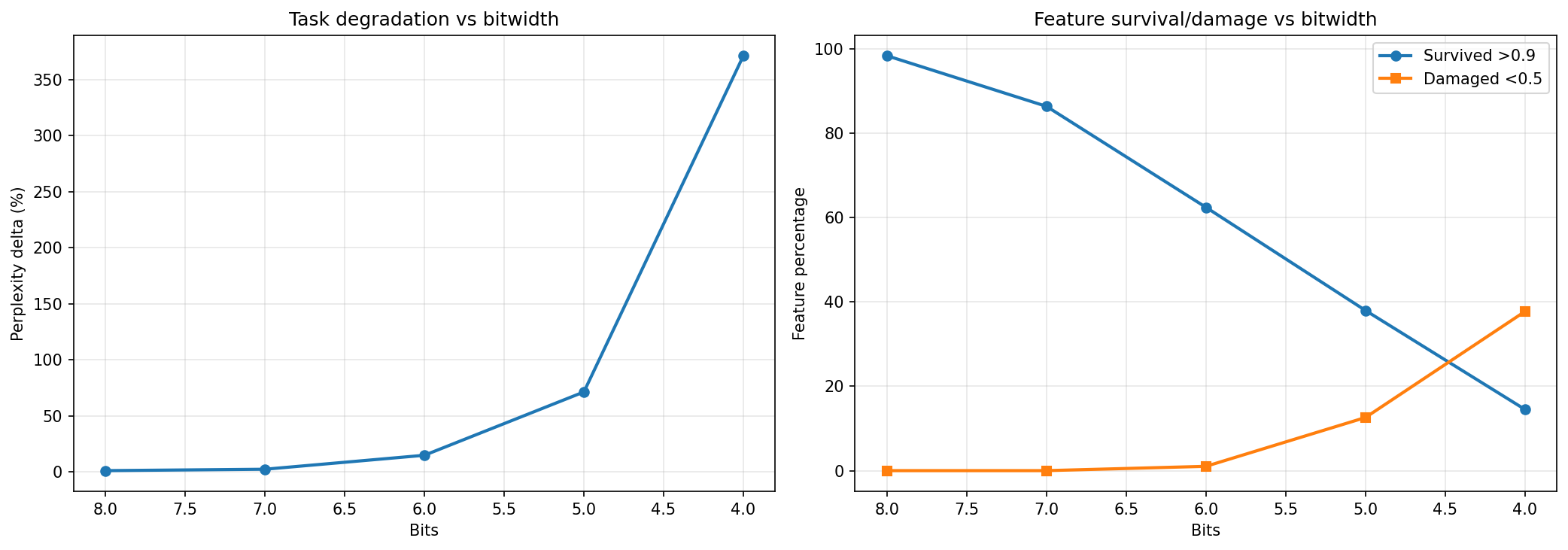}\\[-0.5em]
        {\small (a) \pythiamodel{}}
    \end{minipage}\hfill
    \begin{minipage}{0.49\linewidth}
        \centering
        \includegraphics[width=\linewidth]{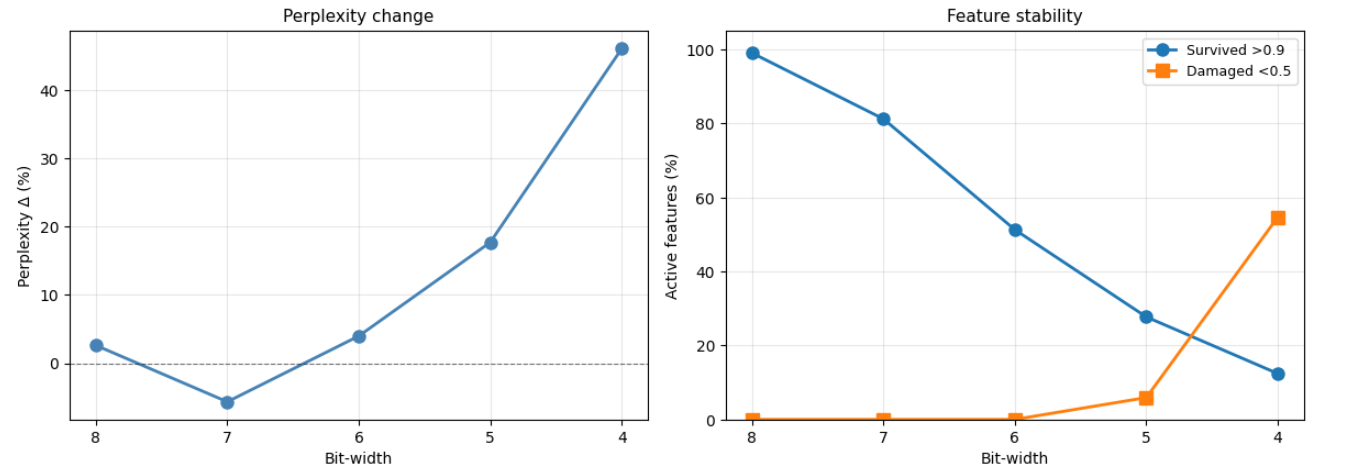}\\[-0.5em]
        {\small (b) \gemmamodel{}}
    \end{minipage}
    \caption{Cross-scale RTN bit-width sweep. Panel (a) shows \pythiamodel{}; panel (b) shows \gemmamodel{}.}\label{fig:cross-model-rtn-sweep}
\end{figure}

Table~\ref{tab:cross-model-rtn-sweep-pruning} reports the round-to-nearest (RTN) bit-width sweep for both models together with a magnitude-pruning baseline; Figure~\ref{fig:cross-model-rtn-sweep} plots survival and damage against bit-width.

\begin{table}[t]
    \centering
    \scriptsize
    \setlength{\tabcolsep}{3pt}
    \caption{Cross-model RTN quantization sweep and pruning baseline.}\label{tab:cross-model-rtn-sweep-pruning}
    \resizebox{\linewidth}{!}{%
    \begin{tabular}{llrrrrrrr}
        \toprule
        Model & Condition & Bits & Prune sparsity & $\Delta$PPL (\%) & Mean corr. & Survived (\%) & Degraded (\%) & Damaged (\%) \\
        \midrule
        \pythiamodel{} & FP16 baseline & 16 & \textemdash{} & 0.000 & 1.000 & 100.000 & 0.000 & 0.000 \\
        \pythiamodel{} & RTN INT8 & 8 & \textemdash{} & 1.020 & 0.981 & 98.312 & 1.688 & 0.000 \\
        \pythiamodel{} & RTN INT7 & 7 & \textemdash{} & 2.258 & 0.957 & 86.322 & 13.678 & 0.000 \\
        \pythiamodel{} & RTN INT6 & 6 & \textemdash{} & 14.722 & 0.888 & 62.395 & 36.586 & 1.020 \\
        \pythiamodel{} & RTN INT5 & 5 & \textemdash{} & 71.208 & 0.777 & 37.957 & 49.455 & 12.588 \\
        \pythiamodel{} & RTN INT4 & 4 & \textemdash{} & 371.112 & 0.583 & 14.504 & 47.785 & 37.711 \\
        \pythiamodel{} & Magnitude pruning matched INT6 & 16 & 0.175 & 13.488 & 0.891 & 60.970 & 38.731 & 0.299 \\
        \midrule
        \gemmamodel{} & FP16 baseline & 16 & \textemdash{} & 0.000 & 1.000 & 100.000 & 0.000 & 0.000 \\
        \gemmamodel{} & RTN INT8 & 8 & \textemdash{} & 2.624 & 0.966 & 99.090 & 0.910 & 0.000 \\
        \gemmamodel{} & RTN INT7 & 7 & \textemdash{} & -5.648 & 0.943 & 81.257 & 18.743 & 0.000 \\
        \gemmamodel{} & RTN INT6 & 6 & \textemdash{} & 3.989 & 0.891 & 51.302 & 48.698 & 0.000 \\
        \gemmamodel{} & RTN INT5 & 5 & \textemdash{} & 17.716 & 0.764 & 27.716 & 66.363 & 5.921 \\
        \gemmamodel{} & RTN INT4 & 4 & \textemdash{} & 46.081 & 0.494 & 12.402 & 32.895 & 54.703 \\
        \gemmamodel{} & Magnitude pruning matched INT6 & \textemdash{} & 0.162 & 6.126 & 0.841 & 38.578 & 61.184 & 0.238 \\
        \bottomrule
    \end{tabular}%
    }
\end{table}

On \pythiamodel{}, feature survival declines monotonically as precision decreases: 98.3\% of active features survive at INT8, 86.3\% at INT7, 62.4\% at INT6, 38.0\% at INT5, and 14.5\% at INT4. The fraction of damaged features remains negligible through INT7 (0.0\%), reaches 1.0\% at INT6, and then rises sharply to 12.6\% at INT5 and 37.7\% at INT4. Mean per-feature correlation falls correspondingly from 0.981 (INT8) to 0.583 (INT4).

\gemmamodel{} exhibits the same qualitative trajectory at a $30\times$ larger parameter count. Survival declines from 99.1\% (INT8) to 81.3\% (INT7), 51.3\% (INT6), 27.7\% (INT5), and 12.4\% (INT4). Damage is again negligible until low bit-widths, remaining at 0.0\% through INT6, rising to 5.9\% at INT5, and reaching 54.7\% at INT4. The two models differ in detail---Gemma retains marginally more features at INT8 and fewer at INT6---but the shape of the curve, a slow decline at high precision followed by an accelerating collapse below INT6, is shared (Figure~\ref{fig:cross-model-rtn-sweep}).

Two features of the sweep are notable for the analysis that follows. First, in both models the transition from ``mostly intact'' to ``mostly degraded'' occurs over a narrow band between INT7 and INT5, rather than as a smooth linear decline. Second, the degraded band (correlation in $[0.5,0.9]$) is consistently larger than the damaged band at intermediate bit-widths: at Gemma INT6, 48.7\% of features are degraded but 0.0\% are damaged, indicating that intermediate quantization predominantly blurs features rather than destroying them.

\subsection{Feature survival is predictable from full-precision statistics}\label{subsec:feature-survival-predictable}

\begin{figure}[H]
    \centering
    \includegraphics[width=\linewidth]{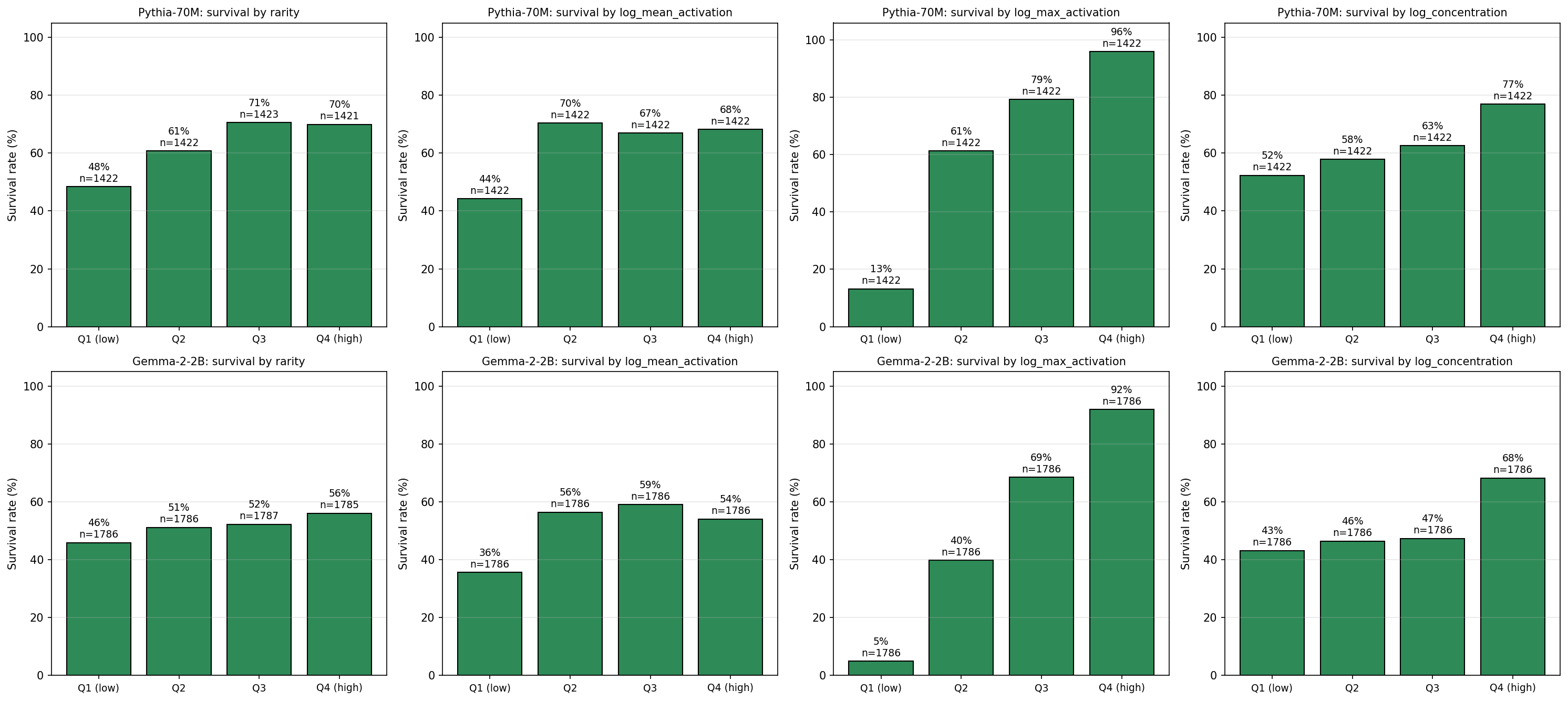}
    \caption{Feature survival by quartile of FP16 feature statistics under RTN INT6. Quartiles are computed within each model and statistic; bars show the percentage of active features with correlation $c_j>0.9$ after quantization.}\label{fig:survival-by-quartile-main}
\end{figure}

We test whether INT6 feature survival can be predicted from FP16 statistics alone. For each active feature, we compute rarity, log mean activation, log peak activation, and log activation concentration, then fit an L2-regularized logistic regression to the survival label $c_j>0.9$. Table~\ref{tab:fragility-predictor-summary} reports coefficients and 5-fold cross-validated AUC; Figure~\ref{fig:survival-by-quartile-main} shows survival by predictor quartile.

\begin{table}[H]
    \centering
    \scriptsize
    \setlength{\tabcolsep}{4pt}
    \caption{Predicting INT6 feature survival from FP16 feature statistics. AUC is the 5-fold cross-validated mean $\pm$ standard deviation; coefficients are standardized logistic-regression coefficients.}\label{tab:fragility-predictor-summary}
    \resizebox{\linewidth}{!}{%
    \begin{tabular}{lrrcrrrr}
        \toprule
        Model & Active feats. & Survived (\%) & AUC & Rarity & Log mean & Log peak & Log concentration \\
        \midrule
        \pythiamodel{} & 5,688 & 62.4 & $0.924 \pm 0.007$ & 4.52 & 2.72 & 1.66 & -1.67 \\
        \gemmamodel{} & 7,144 & 51.3 & $0.971 \pm 0.002$ & 17.84 & 9.79 & 4.14 & -8.16 \\
        \bottomrule
    \end{tabular}%
    }
\end{table}

The predictor is highly accurate in both models: cross-validated AUC is $0.924\pm0.007$ on \pythiamodel{} ($n=5{,}688$) and $0.971\pm0.002$ on \gemmamodel{} ($n=7{,}144$). All four predictors agree in sign across models, suggesting that the relationship between FP16 feature statistics and quantization survival generalizes across scale.

Peak activation is the strongest marginal predictor. Survival rises from 13\% to 96\% across peak-activation quartiles on \pythiamodel{} and from 5\% to 92\% on \gemmamodel{} (Figure~\ref{fig:survival-by-quartile-main}). Rarity has the largest multivariate coefficient (4.52 on \pythiamodel{}, 17.84 on \gemmamodel{}), but because rarity and the activation-magnitude predictors are correlated, the multivariate coefficient ranking is not a reliable guide to which predictor matters most marginally. The robust conclusion is that survival is highly predictable from simple FP16 statistics, peak activation is the strongest single marginal predictor, and rarer features are not more fragile after controlling for activation statistics.

\subsection{Feature damage can diverge from task-level perplexity}\label{subsec:feature-damage-perplexity-divergence}

Table~\ref{tab:cross-model-rtn-sweep-pruning} shows a dissociation between perplexity and feature survival on \gemmamodel{}. INT7 improves perplexity by 5.65\% while survival falls to 81.3\%, meaning 18.7\% of active features degrade despite improved task-level performance. INT6 further reduces feature survival to 51.3\%, even though its perplexity increase remains modest at 3.99\%.

\begin{table}[H]
    \centering
    \scriptsize
    \setlength{\tabcolsep}{5pt}
    \caption{Gemma behavioral protocol check. Chunked and sliding-window perplexity for FP16 and RTN INT8/INT7/INT6, with SAE survival from the main QDM run.}\label{tab:gemma-behavioral-protocol-check}
    \resizebox{\linewidth}{!}{%
    \begin{tabular}{lrrrrr}
        \toprule
        Condition & Chunked PPL & Chunked $\Delta$PPL (\%) & Sliding PPL & Sliding $\Delta$PPL (\%) & SAE survival (\%) \\
        \midrule
        FP16 baseline & 458.53 & 0.00 & 46.39 & 0.00 & 100.00 \\
        RTN INT8 & 470.56 & +2.62 & 46.78 & +0.82 & 99.09 \\
        RTN INT7 & 432.64 & -5.65 & 43.38 & -6.49 & 81.26 \\
        RTN INT6 & 476.82 & +3.99 & 44.79 & -3.45 & 51.30 \\
        \bottomrule
    \end{tabular}%
    }
\end{table}

To check that this dissociation is not an artifact of chunked perplexity, we re-evaluate FP16, INT8, INT7, and INT6 with a sliding-window protocol (window 2048, stride 512; Table~\ref{tab:gemma-behavioral-protocol-check}, \ref{app:sliding-window}). Sliding-window evaluation lowers absolute perplexity but preserves the dissociation: INT7 improves perplexity under both protocols, and INT6 improves perplexity under sliding-window while only 51.3\% of active features survive. The INT7 diagnostic suite in \ref{app:int7-investigation-checks} confirms that the effect is reproducible and not a quantization failure.

\subsection{Quantization and pruning damage overlapping feature sets}\label{subsec:quantization-pruning-overlap}

Table~\ref{tab:rtn-pruning-overlap} compares RTN INT6 against magnitude pruning calibrated to match INT6 perplexity (Pythia sparsity 0.175, perplexity 76.99 vs. INT6 77.83; Gemma sparsity 0.1625, perplexity 486.6 vs. INT6 476.8; calibration in \ref{app:pruning-calibration}). On \pythiamodel{} the two methods produce similar aggregate damage (37.6\% non-survived under RTN, 39.0\% under pruning); on \gemmamodel{} pruning is more aggressive at the matched operating point (48.7\% vs. 61.4\% non-survived).

\begin{table}[H]
    \centering
    \scriptsize
    \setlength{\tabcolsep}{4pt}
    \caption{RTN INT6 versus pruning overlap. Summary of non-survived feature-set overlap and per-feature damage-score correlation between RTN INT6 and calibrated magnitude pruning.}\label{tab:rtn-pruning-overlap}
    \resizebox{\linewidth}{!}{%
    \begin{tabular}{lrrrrr}
        \toprule
        Model & RTN non-survived (\%) & Pruning non-survived (\%) & Jaccard overlap & Pearson damage corr. & Spearman damage corr. \\
        \midrule
        \pythiamodel{} & 37.61 & 39.03 & 0.860 & 0.951 & 0.976 \\
        \gemmamodel{} & 48.70 & 61.42 & 0.792 & 0.959 & 0.978 \\
        \bottomrule
    \end{tabular}%
    }
\end{table}

At the per-feature level the two methods are strongly concordant. The Jaccard overlap of non-survived features is 0.86 (Pythia) and 0.79 (Gemma), and the Spearman correlation of per-feature damage scores is 0.98 in both models (Table~\ref{tab:rtn-pruning-overlap}; scatter and decile plots in \ref{app:rtn-versus-pruning-overlap-visualizations}). The methods differ in the tail: RTN produces no damaged ($c_j<0.5$) features on Gemma while pruning produces 17, and the firing-rate decile analysis (\ref{app:rtn-versus-pruning-overlap-visualizations}) shows pruning damaging more features than RTN in every decile on Gemma, with the gap largest at high firing rates. The decile analysis also shows, for both methods and both models, that non-survival rises with firing rate---high-firing features are more vulnerable---consistent with Section~\ref{subsec:feature-survival-predictable}.

\subsection{Methodology ablations show stable QDM estimates}\label{subsec:methodology-ablations}

\begin{figure}[H]
    \centering
    \includegraphics[width=\linewidth]{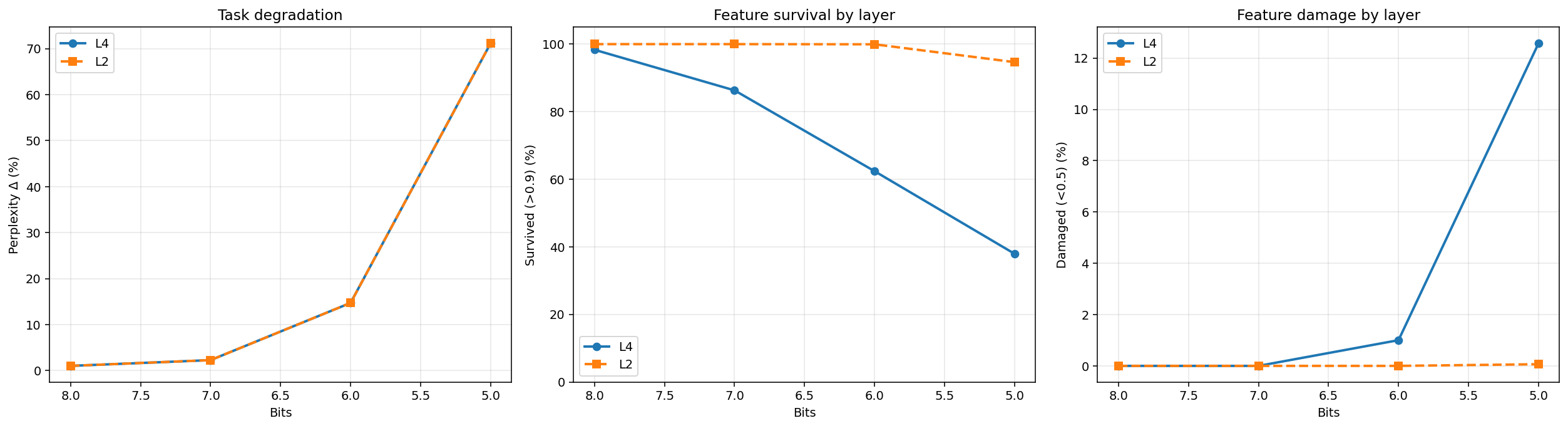}
    \caption{Layer-sensitivity check for \pythiamodel{}. RTN bit-width sweep repeated at two residual-stream layers, comparing feature survival across layers.}\label{fig:pythia-layer-comparison}
\end{figure}

Table~\ref{tab:methodology-ablations-pythia-int6} summarizes methodology ablations for \pythiamodel{} INT6. QDM estimates are stable across token budgets and random subsets: survival is 60.7--62.8\% across 50k--200k tokens, and three independent 100k-token subsets vary by only 0.71 percentage points. The FP16-vs-FP16 null gives mean correlation 1.000000 and 0.0\% damaged features, confirming that the pipeline does not manufacture drift. Threshold sensitivity checks preserve the qualitative pattern (\ref{app:stability-ablation-threshold-sensitivity}), and the bit-width sweep also holds at a second layer (Figure~\ref{fig:pythia-layer-comparison}), indicating that the result is not specific to the primary read-out site.

\begin{table}[H]
    \centering
    \small
    \caption{Methodology ablations for \pythiamodel{} INT6. Summary of token-budget, random-subset, FP16-null, and threshold-sensitivity checks used to assess QDM measurement stability.}\label{tab:methodology-ablations-pythia-int6}
    \begin{tabular}{ccccc}
        \toprule
        Survival threshold & Damage threshold & Survived & Degraded & Damaged \\
        \midrule
        0.80 & 0.50 & 80.13\% & 18.85\% & 1.02\% \\
        0.90 & 0.50 & 62.39\% & 36.59\% & 1.02\% \\
        0.95 & 0.50 & 44.41\% & 54.57\% & 1.02\% \\
        \bottomrule
    \end{tabular}
\end{table}

\section{Analysis and Discussion}\label{sec:analysis-discussion}

\subsection{Task metrics are insufficient indicators of feature-level fidelity}\label{subsec:discussion-task-metrics-insufficient}

The most consequential dissociation is between perplexity and feature survival (Section~\ref{subsec:feature-damage-perplexity-divergence}). At Gemma INT7, perplexity improves while nearly one in five features degrades; under sliding-window evaluation INT6 also improves perplexity while roughly half of features fall below the survival threshold. The INT6 sign change between the chunked and sliding-window protocols underscores the point: the task metric is sensitive enough to the evaluation protocol that its sign can flip, while the feature-survival measurement---computed on fixed tokens---does not depend on the perplexity protocol at all.

One plausible interpretation is that INT7 rounding acts as mild regularization under this corpus and protocol, but we do not establish the mechanism at a bit-width where the perturbation is large enough to be beneficial but not yet destructive; the logit-drift diagnostics (\ref{app:int7-investigation-checks}) show INT7 perturbs the model's outputs more than INT8 and less than INT6, yet reduces loss where INT8 and INT6 do not. The mechanism is secondary to the implication: a practitioner selecting a quantization configuration by perplexity alone could choose a setting that appears lossless, or even beneficial, while having substantially altered the model's internal feature structure. For interpretability work that is developed on a full-precision model and then deployed in quantized form, perplexity parity is therefore not sufficient evidence that the analyzed features remain intact.

\subsection{Vulnerability is shared across compression methods}\label{subsec:discussion-shared-vulnerability}

The matched-perplexity comparison (Section~\ref{subsec:quantization-pruning-overlap}) shows that quantization and pruning, at equal task degradation, damage strongly overlapping sets of features---Jaccard 0.79 to 0.86 and damage-score Spearman 0.98. The two methods are therefore better described as agreeing on which features are vulnerable than as targeting different feature classes. They differ in severity: on Gemma, matched pruning is somewhat more aggressive overall (and was calibrated to a slightly higher perplexity than INT6, so part of this gap reflects residual mismatch), and pruning produces a small tail of fully damaged features that quantization does not. 

The shared vulnerability ranking connects directly to the signal-strength account in Section~\ref{subsec:discussion-fragility-predictable}: if feature survival depends on whether a feature's signal exceeds the perturbation scale, then both rounding and removal should endanger many of the same low-signal features. The high cross-method damage-score correlation matches this prediction and extends the pruning-based picture of \citet{borobia2026pruning} to quantization: feature vulnerability also appears under weight rounding, where it can be traced across bit-widths, compared to perplexity, and measured on the same frozen SAE basis.

\subsection{Quantization induces graded mechanistic change, not a precision cliff}\label{subsec:discussion-graded-mechanistic-change}

Feature survival declines systematically with bit-width rather than remaining intact until a sharp precision floor. Because correlations are computed on identical tokens with a fixed SAE, this isolates the effect of weight rounding on the residual-stream geometry read out by the SAE. The same pattern across \pythiamodel{} and \gemmamodel{} suggests that quantization fragility is not specific to one model or scale.

At intermediate bit-widths, degradation dominates damage (e.g., Gemma INT6: 48.7\% degraded, 0\% damaged): features are often blurred rather than destroyed. Their activations remain positively correlated with the full-precision feature, but no longer strongly enough to be treated as the same interpretability unit. This matters because SAE analyses typically assume features are stable referents; a feature at correlation 0.7 is neither the original feature nor pure noise, and analyses transferred from FP16 to a quantized model can silently inherit this blurring.

\subsection{Fragility is structured and predictable}\label{subsec:discussion-fragility-predictable}

The logistic-regression results (Section~\ref{subsec:feature-survival-predictable}) show that feature survival is highly predictable from simple FP16 statistics. Mechanistically, quantization perturbations compete with each feature's activation signal: high-signal features remain above the perturbation floor, while weak-signal features are more easily blurred. This explains why peak activation is the strongest marginal predictor and why the predictors generalize across model scales.

The role of rarity requires care. Rarity has the largest standardized coefficient in the multivariate model, and its sign is positive---rarer features survive better. This is initially counter-intuitive if one expects specialized, rarely firing features to be fragile. However, the marginal effect of rarity is modest relative to peak activation, and the two predictors are correlated, so the large multivariate coefficient should not be read as rarity being the dominant causal factor. The defensible statement is that survival is jointly predictable from these statistics with high accuracy, that peak activation is the strongest single marginal predictor, and that, controlling for the other variables, rarer features are not more fragile and if anything survive better.

\citet{borobia2026pruning} report the same pattern under unstructured pruning. Our results show the pattern is not specific to weight removal: it also holds under weight rounding. The agreement across two mechanically distinct compression operations strengthens the interpretation that vulnerability is governed by a feature's signal strength rather than by the specific way weights are altered.

\section{Conclusion}
We asked whether behavioral parity under quantization implies SAE feature fidelity. Holding the SAE, token set, and read-out site fixed while varying only model weights, we measured per-feature survival across RTN bit-widths on Pythia-70M and Gemma-2-2B. Three findings stand out. First, task-level metrics can miss feature damage: perplexity can remain stable or improve while many SAE features degrade. Second, quantization and matched-perplexity pruning damage many of the same features, suggesting a shared compression-induced vulnerability. Third, feature survival under quantization is graded and predictable: features are often blurred rather than destroyed, and survival can be forecast from full-precision statistics alone, with peak activation as the strongest marginal predictor.

These results show that behavioral parity is not sufficient evidence that full-precision interpretability transfers to quantized deployment. An analysis and a quantization setting can each appear sound in isolation while failing to compose. We release our pipeline and per-feature results, and view causal validation as the next step: testing whether these feature changes alter steering, auditing, or other safety-relevant interventions in deployed quantized models.

\section{Limitations}\label{sec:limitations}

Our study is limited to two model families, publicly available residual-stream SAEs, and simulated round-to-nearest weight quantization. We evaluate fixed token budgets, read-out layers, and compression settings, so the quantitative survival rates should not be interpreted as universal across architectures, datasets, SAE training recipes, activation sites, or deployment quantizers. We also focus on feature-level correlation and perplexity rather than downstream tasks or causal interventions; future work should test whether the same feature changes alter steering, auditing, or safety-relevant behavior in deployed quantized models.

\begingroup
\raggedright
\bibliography{main}
\bibliographystyle{tmlr}
\endgroup

\clearpage
\appendix
\renewcommand{\thesection}{Appendix \Alph{section}}
\renewcommand{\theHsection}{appendix.\Alph{section}}
\section{Full all-condition compression plots}\label{app:full-all-condition-compression-plots}
\renewcommand{\thefigure}{A.\arabic{figure}}
\renewcommand{\theHfigure}{A.\arabic{figure}}
\renewcommand{\figurename}{Appendix Figure}
\setcounter{figure}{0}

Appendix A reports the full all-condition bar plots for \pythiamodel{} and \gemmamodel{}. These figures supplement the main RTN bit-width sweep by showing all evaluated compression conditions, including FP16, RTN INT8--INT4, and the pruning baseline.

\begin{figure}[H]
    \centering
    \includegraphics[width=0.95\linewidth]{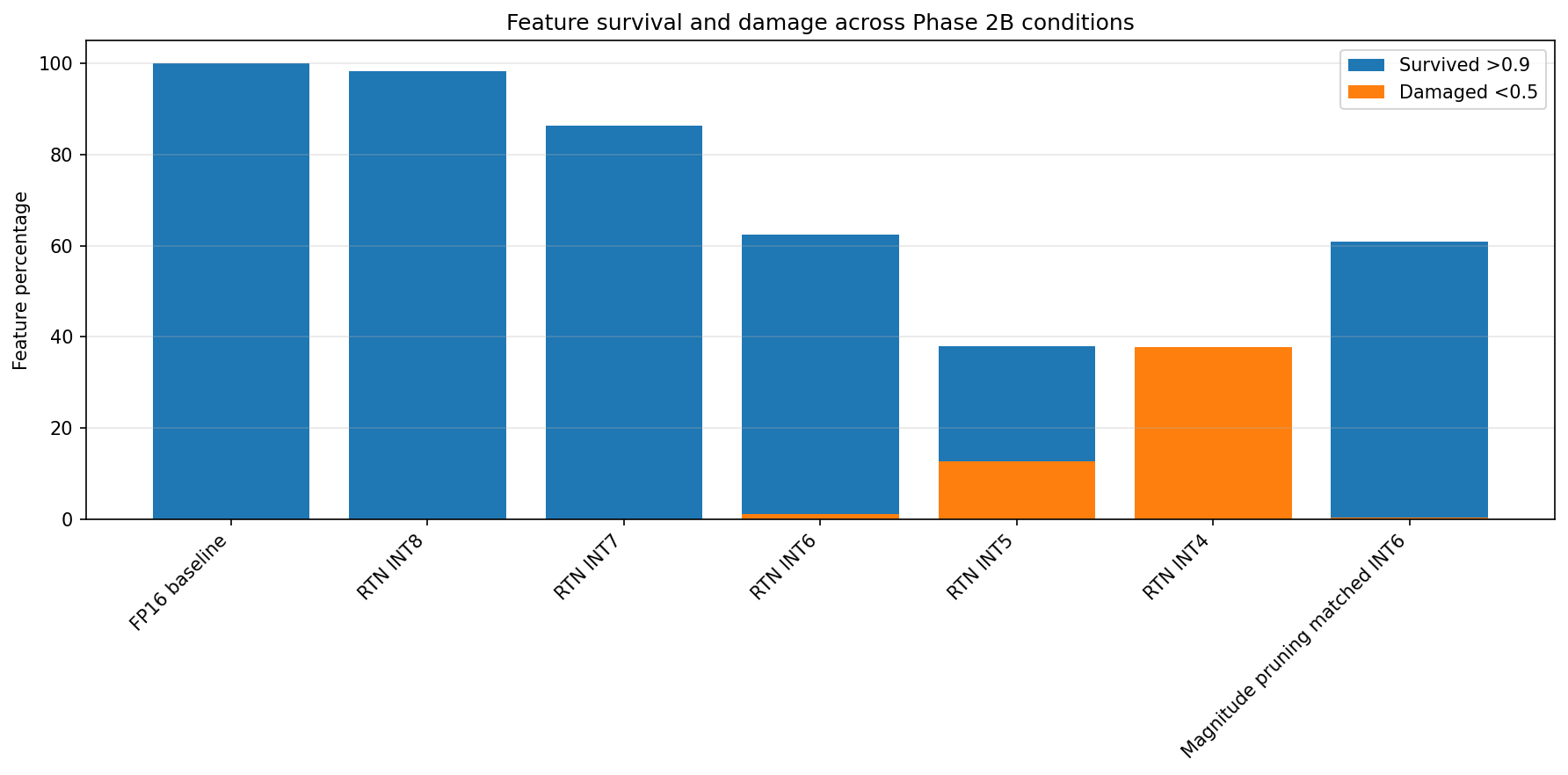}
    \caption{Full \pythiamodel{} compression comparison across FP16, RTN bit-widths, and matched pruning.}\label{fig:phase2b-all-conditions-bar}
\end{figure}

\begin{figure}[H]
    \centering
    \includegraphics[width=0.95\linewidth]{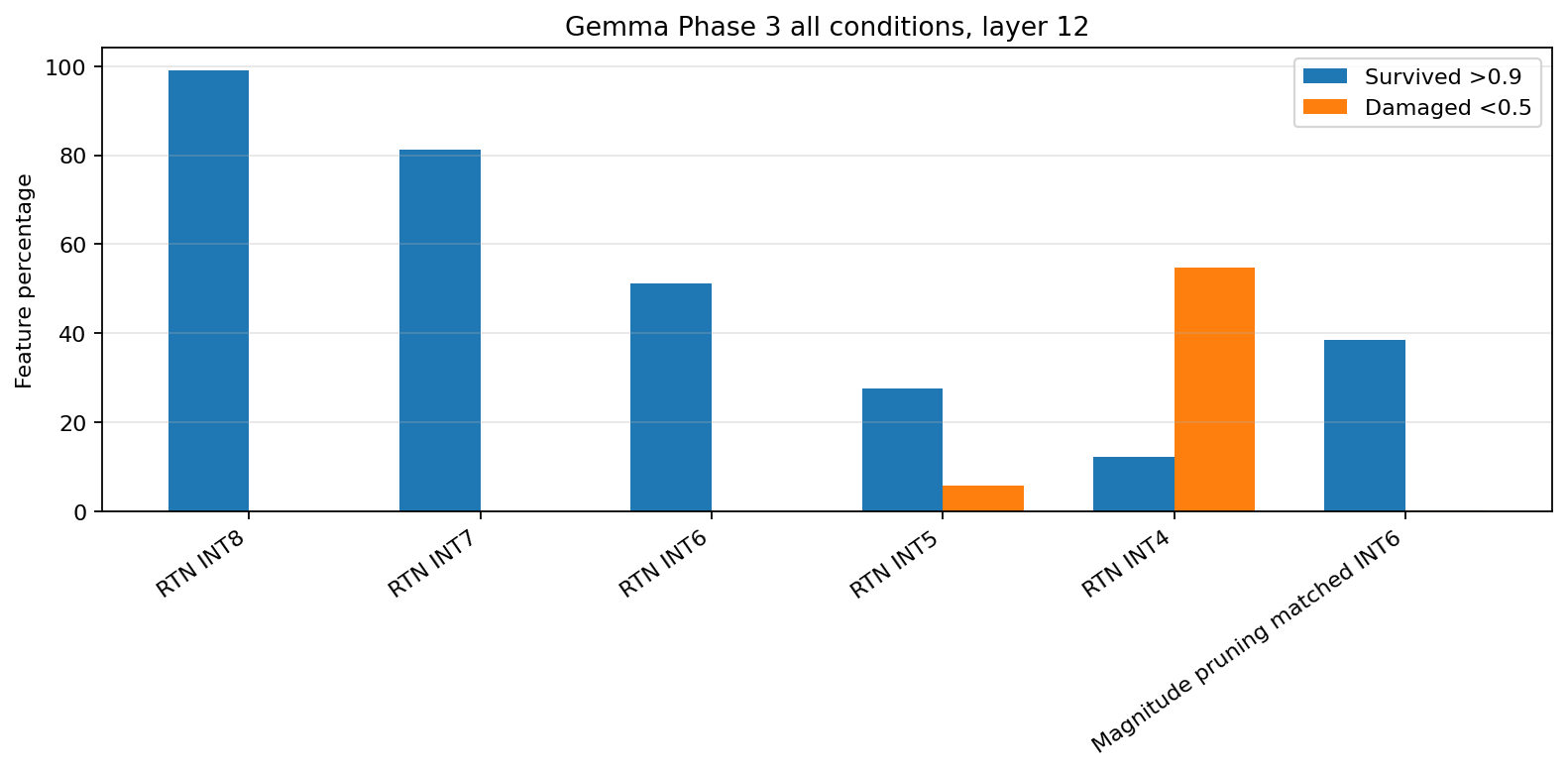}
    \caption{Full \gemmamodel{} compression comparison across FP16, RTN bit-widths, and approximately matched pruning.}\label{fig:phase3-all-conditions-bar}
\end{figure}

\clearpage
\setcounter{section}{1}
\section{Stability ablation plots and threshold sensitivity}\label{app:stability-ablation-threshold-sensitivity}
\renewcommand{\thefigure}{B.\arabic{figure}}
\renewcommand{\thetable}{B.\arabic{table}}
\renewcommand{\theHfigure}{B.\arabic{figure}}
\renewcommand{\theHtable}{B.\arabic{table}}
\renewcommand{\tablename}{Appendix Table}
\setcounter{figure}{0}
\setcounter{table}{0}

\begin{figure}[htbp]
    \centering
    \includegraphics[width=\linewidth]{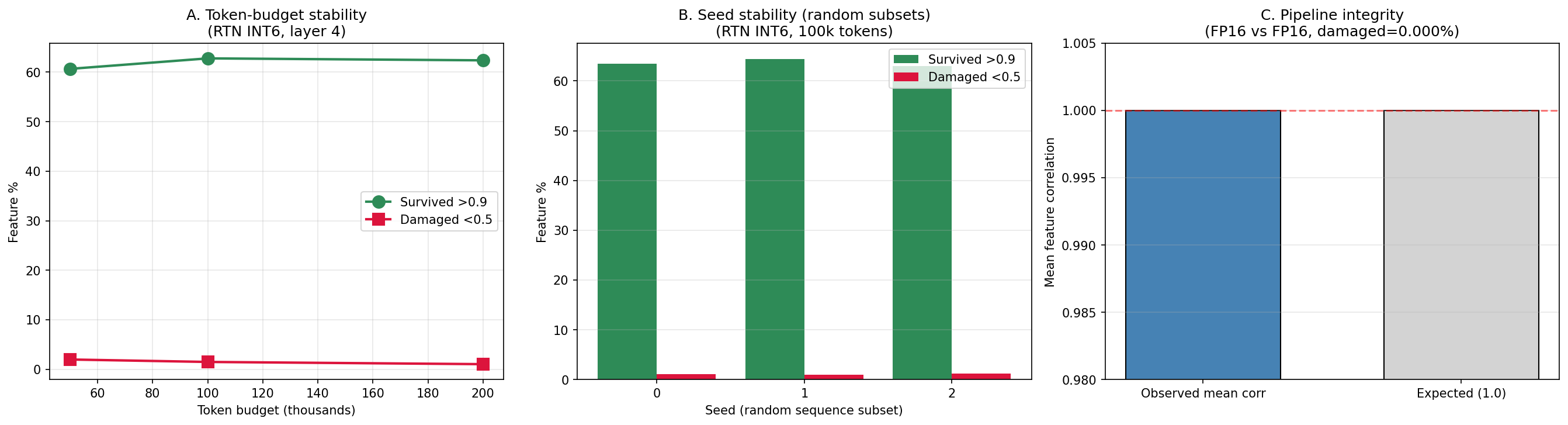}
    \caption{Stability ablation plots for Phase 4.}\label{fig:phase4-stability-ablations}
\end{figure}

\begin{table}[htbp]
    \centering
    \small
    \caption{Threshold sensitivity for ablation D. Percentages are reported over all evaluated conditions.}\label{tab:ablation-d-threshold-sensitivity}
    \resizebox{\linewidth}{!}{%
    \begin{tabular}{ccccc}
        \toprule
        Survival threshold & Damage threshold & Survived (\%) & Degraded (\%) & Damaged (\%) \\
        \midrule
        0.80 & 0.30 & 80.13 & 19.83 & 0.04 \\
        0.80 & 0.40 & 80.13 & 19.64 & 0.23 \\
        0.80 & 0.50 & 80.13 & 18.85 & 1.02 \\
        0.80 & 0.60 & 80.13 & 15.84 & 4.03 \\
        0.80 & 0.70 & 80.13 & 10.36 & 9.51 \\
        0.85 & 0.30 & 73.12 & 26.85 & 0.04 \\
        0.85 & 0.40 & 73.12 & 26.65 & 0.23 \\
        0.85 & 0.50 & 73.12 & 25.86 & 1.02 \\
        0.85 & 0.60 & 73.12 & 22.86 & 4.03 \\
        0.85 & 0.70 & 73.12 & 17.37 & 9.51 \\
        0.90 & 0.30 & 62.39 & 37.57 & 0.04 \\
        0.90 & 0.40 & 62.39 & 37.38 & 0.23 \\
        0.90 & 0.50 & 62.39 & 36.59 & 1.02 \\
        0.90 & 0.60 & 62.39 & 33.58 & 4.03 \\
        0.90 & 0.70 & 62.39 & 28.09 & 9.51 \\
        0.95 & 0.30 & 44.41 & 55.56 & 0.04 \\
        0.95 & 0.40 & 44.41 & 55.36 & 0.23 \\
        0.95 & 0.50 & 44.41 & 54.57 & 1.02 \\
        0.95 & 0.60 & 44.41 & 51.56 & 4.03 \\
        0.95 & 0.70 & 44.41 & 46.08 & 9.51 \\
        \bottomrule
    \end{tabular}%
    }
\end{table}

\begin{table}[htbp]
    \centering
    \small
    \caption{Token-budget stability ablation. This table reports QDM feature-survival statistics for RTN INT6 using increasing token budgets. It tests whether estimated survival and damage rates are sensitive to the number of evaluated tokens.}\label{tab:ablation-a-token-budget}
    \resizebox{\linewidth}{!}{%
    \begin{tabular}{cccccc}
        \toprule
        Token budget & Evaluated tokens & Active features & Mean corr. & Survived (\%) & Damaged (\%) \\
        \midrule
        50k & 49,664 & 5,609 & 0.8796 & 60.67 & 1.96 \\
        100k & 99,840 & 5,719 & 0.8858 & 62.79 & 1.47 \\
        200k & 199,680 & 5,688 & 0.8879 & 62.39 & 1.02 \\
        \bottomrule
    \end{tabular}%
    }
\end{table}

\begin{table}[H]
    \centering
    \small
    \caption{Random-subset seed stability ablation. This table reports QDM statistics across independently sampled 100k-token subsets. It tests whether the measured feature-survival rate depends on the random token subset used for activation comparison.}\label{tab:ablation-b-seed-stability}
    \resizebox{\linewidth}{!}{%
    \begin{tabular}{cccccc}
        \toprule
        Seed & Evaluated tokens & Active features & Mean corr. & Survived (\%) & Damaged (\%) \\
        \midrule
        0 & 99,840 & 5,721 & 0.8905 & 63.52 & 1.01 \\
        1 & 99,840 & 5,788 & 0.8926 & 64.39 & 0.93 \\
        2 & 99,840 & 5,726 & 0.8883 & 62.98 & 1.24 \\
        \bottomrule
    \end{tabular}%
    }
\end{table}

\begin{table}[H]
    \centering
    \small
    \caption{FP16-vs-FP16 pipeline null check. This table compares FP16 activations against the FP16 reference under the same QDM pipeline. The expected result is mean correlation 1.0, 100\% survival, and 0\% damage, confirming that the pipeline does not create artificial feature drift.}\label{tab:ablation-c-random-baseline}
    \resizebox{\linewidth}{!}{%
    \begin{tabular}{lcccc}
        \toprule
        Test & Active features & Mean corr. & Survived (\%) & Damaged (\%) \\
        \midrule
        Strict null (same model, same tokens) & 5,719 & 1.0000 & 100.00 & 0.00 \\
        \bottomrule
    \end{tabular}%
    }
\end{table}

\clearpage
\section{INT7 investigation checks}\label{app:int7-investigation-checks}
\renewcommand{\thetable}{C.\arabic{table}}
\renewcommand{\theHtable}{C.\arabic{table}}
\setcounter{table}{0}

\begin{table}[H]
    \centering
    \small
    \caption{Corrected 50k-token perplexity sweep verifying the INT7 decrease persists under shifted-loss evaluation.}\label{tab:int7-corrected-ppl-sweep}
    \resizebox{\linewidth}{!}{%
    \begin{tabular}{lccccc}
        \toprule
        Condition & Bits & Loss & Perplexity & PPL delta (\%) & Prediction tokens \\
        \midrule
        FP16 & 16 & 6.2134 & 499.40 & 0.00 & 49,725 \\
        RTN INT8 & 8 & 6.2335 & 509.52 & 2.03 & 49,725 \\
        RTN INT7 & 7 & 6.1520 & 469.66 & -5.96 & 49,725 \\
        RTN INT6 & 6 & 6.2481 & 517.04 & 3.53 & 49,725 \\
        \bottomrule
    \end{tabular}%
    }
\end{table}

\begin{table}[H]
    \centering
    \small
    \caption{Five-seed subset check showing the INT7 perplexity decrease is reproducible across token samples.}\label{tab:int7-subset-reproducibility}
    \resizebox{\linewidth}{!}{%
    \begin{tabular}{ccccccccc}
        \toprule
        Seed & Prediction tokens & FP16 PPL & INT8 PPL & INT7 PPL & INT6 PPL & INT8 delta (\%) & INT7 delta (\%) & INT6 delta (\%) \\
        \midrule
        0 & 76,500 & 495.06 & 505.05 & 463.97 & 515.05 & 2.02 & -6.28 & 4.04 \\
        1 & 76,500 & 458.03 & 467.96 & 427.55 & 473.73 & 2.17 & -6.65 & 3.43 \\
        2 & 76,500 & 448.90 & 458.81 & 419.73 & 470.33 & 2.21 & -6.50 & 4.77 \\
        3 & 76,500 & 493.42 & 504.91 & 460.44 & 510.32 & 2.33 & -6.69 & 3.42 \\
        4 & 76,500 & 468.28 & 479.36 & 437.37 & 484.99 & 2.37 & -6.60 & 3.57 \\
        \bottomrule
    \end{tabular}%
    }
\end{table}

\begin{table}[H]
    \centering
    \small
    \caption{Weight-error diagnostics confirming INT7 quantization behaves normally between INT8 and INT6.}\label{tab:int7-weight-diagnostics}
    \resizebox{\linewidth}{!}{%
    \begin{tabular}{ccccccc}
        \toprule
        Bits & Mean MSE & Mean MAE & Mean relative MAE & Mean cosine & Min q seen & Max q seen \\
        \midrule
        4 & $4.51\times10^{-6}$ & 0.001698 & 0.2141 & 0.9805 & -7 & 7 \\
        5 & $9.92\times10^{-7}$ & 0.000794 & 0.1002 & 0.9956 & -15 & 15 \\
        6 & $2.34\times10^{-7}$ & 0.000385 & 0.0485 & 0.9990 & -31 & 31 \\
        7 & $5.74\times10^{-8}$ & 0.000190 & 0.0239 & 0.9997 & -63 & 63 \\
        8 & $1.50\times10^{-8}$ & 0.000095 & 0.0120 & 0.9999 & -128 & 127 \\
        \bottomrule
    \end{tabular}%
    }
\end{table}

\begin{table}[H]
    \centering
    \small
    \caption{Logit-drift check showing INT7 output perturbation lies between INT8 and INT6.}\label{tab:int7-logit-drift}
    \resizebox{\linewidth}{!}{%
    \begin{tabular}{ccccccccc}
        \toprule
        Bits & MSE mean & MSE std. & MAE mean & MAE std. & Cosine mean & Cosine std. & Loss delta mean & Loss delta std. \\
        \midrule
        6 & 1.7596 & 0.4365 & 1.0002 & 0.1072 & 0.9779 & 0.0068 & -0.0012 & 0.0831 \\
        7 & 0.6222 & 0.1735 & 0.5827 & 0.0560 & 0.9922 & 0.0021 & -0.0543 & 0.0391 \\
        8 & 0.1758 & 0.0397 & 0.3104 & 0.0289 & 0.9978 & 0.0006 & 0.0252 & 0.0310 \\
        \bottomrule
    \end{tabular}%
    }
\end{table}

\clearpage
\section{Gemma sliding-window behavioral evaluation}\label{app:sliding-window}\label{app:gemma-sliding-window-behavioral-evaluation}
\renewcommand{\thetable}{D.\arabic{table}}
\renewcommand{\theHtable}{D.\arabic{table}}
\setcounter{table}{0}

To check whether the high absolute Gemma perplexity under the chunked Phase 3 protocol was caused by context resets, we reran behavioral evaluation with a sliding-window protocol. 
This check was applied to FP16, RTN INT8, RTN INT7, and RTN INT6 \gemmamodel{} on WikiText-2-raw.

The evaluation used a HuggingFace-format Gemma model in bfloat16 with per-output-channel RTN quantization. 
The token stream contained $288{,}894$ tokens, of which $288{,}893$ were scored. 
We used window size $W=2048$ and stride $S=512$. 
For each window, previous tokens served as context and only newly introduced target positions were included in the loss. 
The aggregate perplexity was computed as a token-weighted mean negative log-likelihood:
\begin{equation}
\mathrm{PPL}
=
\exp\left(
\frac{\sum_w m_w \mathcal{L}_w}{\sum_w m_w}
\right),
\label{eq:appendix-sliding-window-ppl}
\end{equation}
where $m_w$ is the number of scored tokens in window $w$ and $\mathcal{L}_w$ is the mean loss over those scored tokens.

For the quantized conditions, the HuggingFace RTN implementation quantized $182$ tensors and $2{,}024{,}275{,}968$ parameters. 
This behavioral check is distinct from the main SAE feature-survival pipeline, which uses TransformerLens/Gemma Scope activations. 
We therefore use the sliding-window results only as a robustness check on the perplexity trend, not as a replacement for the main feature-correlation results or as a benchmark-comparable Gemma perplexity evaluation.

\begin{table}[H]
    \centering
    \scriptsize
    \caption{Full Gemma sliding-window perplexity results using window size 2048 and stride 512.}\label{tab:gemma-sliding-window-ppl}
    \resizebox{\linewidth}{!}{%
    \begin{tabular}{lcccccccccc}
        \toprule
        Condition & Bits & Loss & PPL & Total tokens & Loss tokens & Window & Stride & Quantized tensors & Quantized params & PPL delta (\%) \\
        \midrule
        FP16 baseline & 16 & 3.8372 & 46.39 & 288,894 & 288,893 & 2048 & 512 & 0 & 0 & 0.00 \\
        RTN INT8 & 8 & 3.8454 & 46.78 & 288,894 & 288,893 & 2048 & 512 & 182 & 2,024,275,968 & 0.82 \\
        RTN INT7 & 7 & 3.7700 & 43.38 & 288,894 & 288,893 & 2048 & 512 & 182 & 2,024,275,968 & -6.49 \\
        RTN INT6 & 6 & 3.8020 & 44.79 & 288,894 & 288,893 & 2048 & 512 & 182 & 2,024,275,968 & -3.45 \\
        \bottomrule
    \end{tabular}%
    }
\end{table}

\begin{table}[H]
    \centering
    \scriptsize
    \caption{Comparison of chunked and sliding-window Gemma perplexity with corresponding SAE feature-survival rates.}\label{tab:gemma-chunked-vs-sliding-ppl}
    \resizebox{\linewidth}{!}{%
    \begin{tabular}{lccccccccc}
        \toprule
        Condition & Chunked PPL & Chunked delta (\%) & Sliding PPL & Sliding delta (\%) & Survived (\%) & Degraded (\%) & Damaged (\%) & PPL ratio & Median corr. \\
        \midrule
        FP16 baseline & 458.53 & 0.00 & 46.39 & 0.00 & 100.00 & 0.00 & 0.00 & 9.88 & 1.0000 \\
        RTN INT8 & 470.56 & 2.62 & 46.78 & 0.82 & 99.09 & 0.91 & 0.00 & 10.06 & 0.9707 \\
        RTN INT7 & 432.64 & -5.65 & 43.38 & -6.49 & 81.26 & 18.74 & 0.00 & 9.97 & 0.9499 \\
        RTN INT6 & 476.82 & 3.99 & 44.79 & -3.45 & 51.30 & 48.70 & 0.00 & 10.65 & 0.9027 \\
        \bottomrule
    \end{tabular}%
    }
\end{table}

\clearpage
\section{RTN-versus-pruning overlap visualizations}\label{app:rtn-versus-pruning-overlap-visualizations}
\renewcommand{\thefigure}{E.\arabic{figure}}
\renewcommand{\theHfigure}{E.\arabic{figure}}
\renewcommand{\figurename}{Appendix Figure}
\setcounter{figure}{0}

Appendix E visualizes the RTN-versus-pruning overlap analysis. The scatter plots compare per-feature survival correlations under RTN INT6 and magnitude pruning, while the decile plots show how non-survival varies across FP16 firing-rate deciles. These figures support the main-text result that quantization and pruning largely affect overlapping vulnerable feature sets rather than unrelated feature classes.

\begin{figure}[H]
    \centering
    \includegraphics[width=0.95\linewidth]{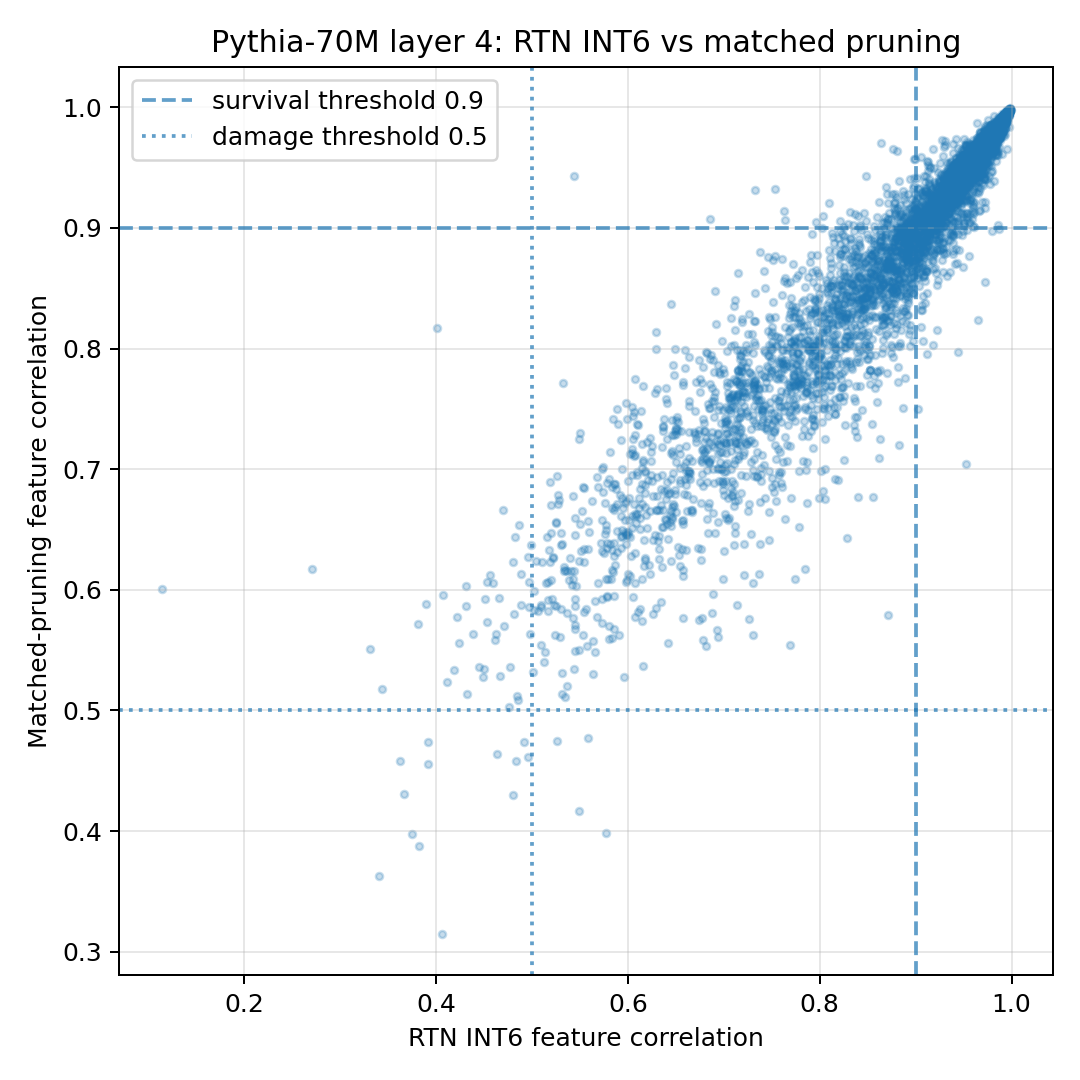}
    \caption{\pythiamodel{} per-feature correlation scatter comparing RTN INT6 and matched magnitude pruning.}\label{fig:phase5b-pythia-rtn-vs-pruning-scatter}
\end{figure}

\begin{figure}[H]
    \centering
    \includegraphics[width=0.95\linewidth]{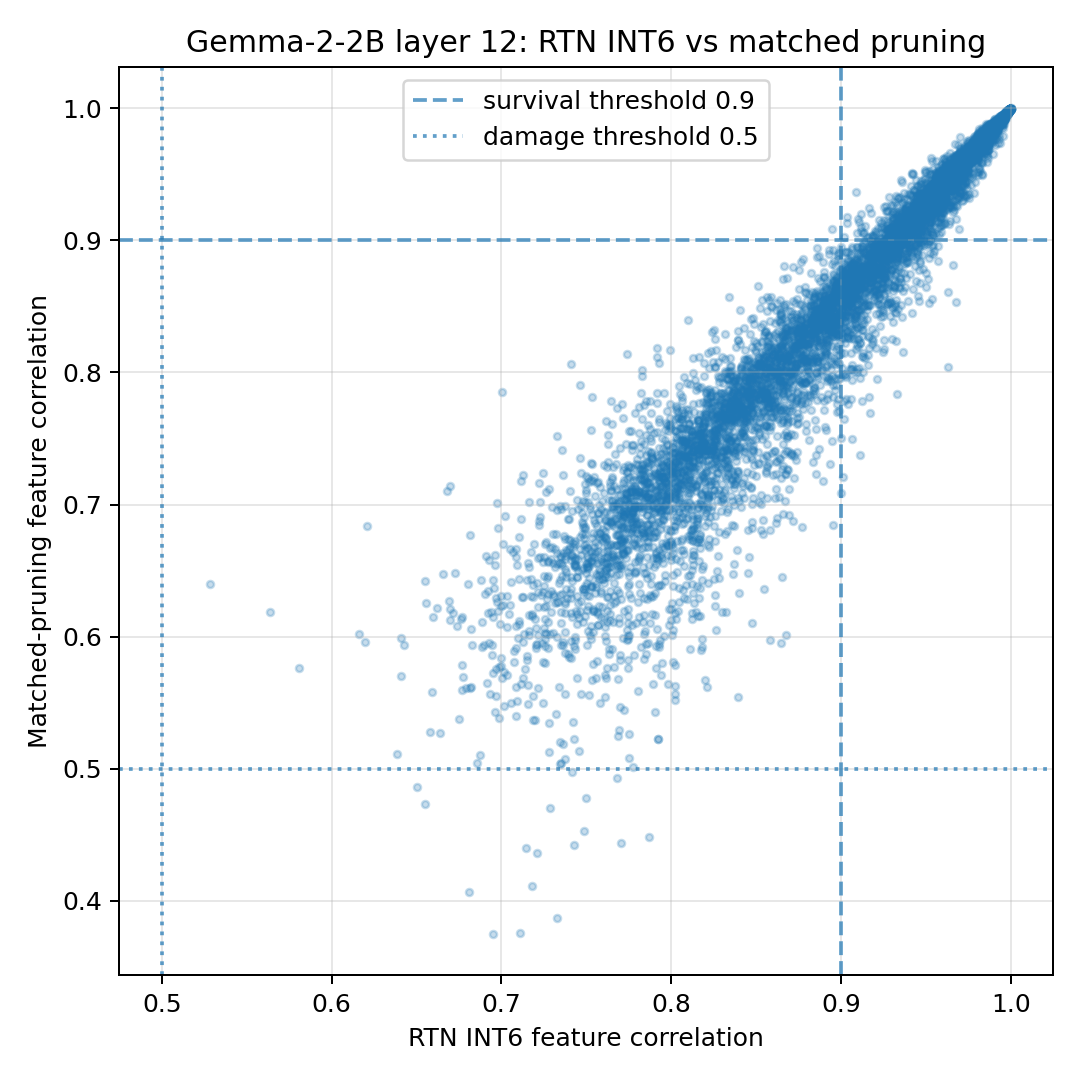}
    \caption{\gemmamodel{} per-feature correlation scatter comparing RTN INT6 and approximately matched magnitude pruning.}\label{fig:phase5b-gemma-rtn-vs-pruning-scatter}
\end{figure}

\begin{figure}[H]
    \centering
    \includegraphics[width=0.95\linewidth]{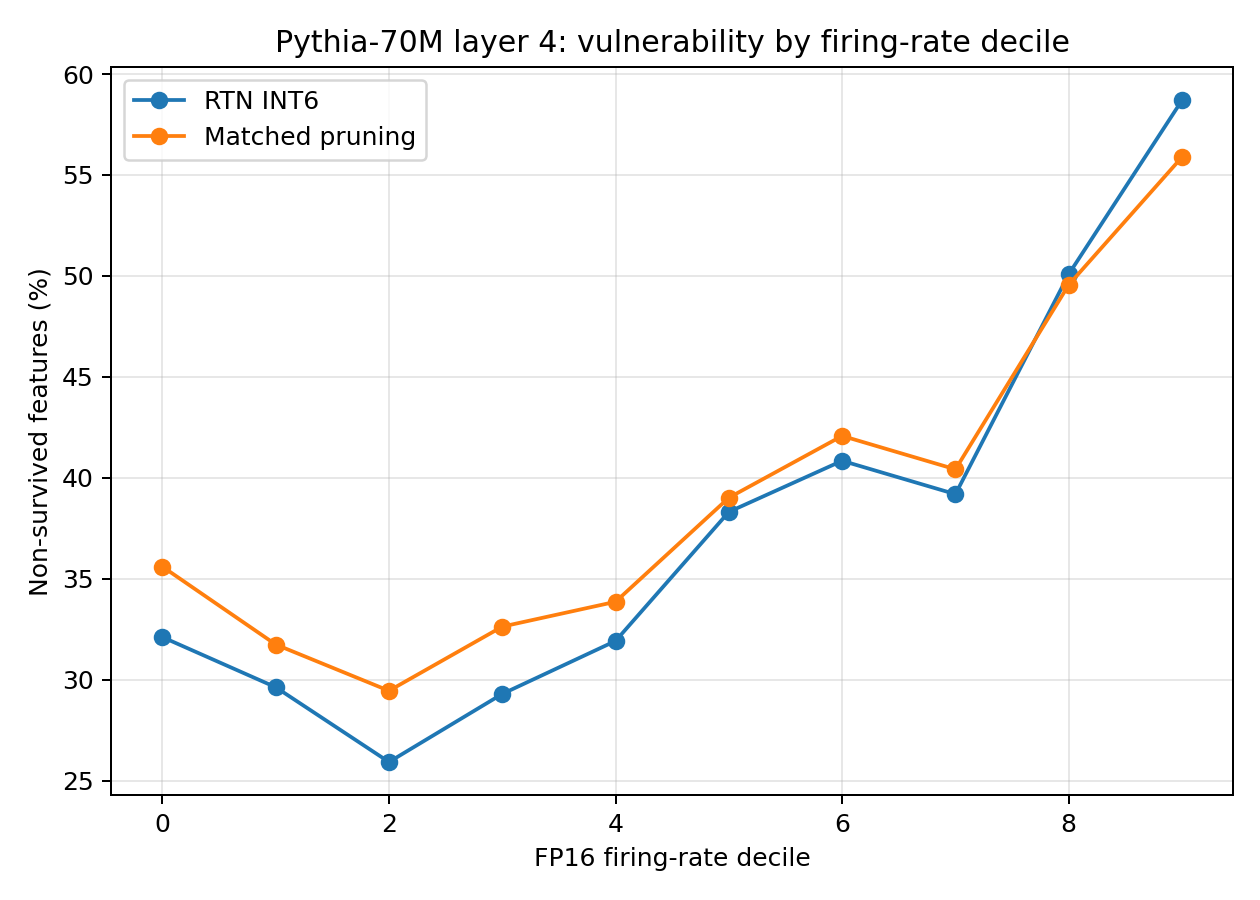}
    \caption{\pythiamodel{} non-survival rates by FP16 firing-rate decile for RTN INT6 and matched pruning.}\label{fig:phase5b-pythia-rtn-vs-pruning-firing-rate-deciles}
\end{figure}

\begin{figure}[H]
    \centering
    \includegraphics[width=0.95\linewidth]{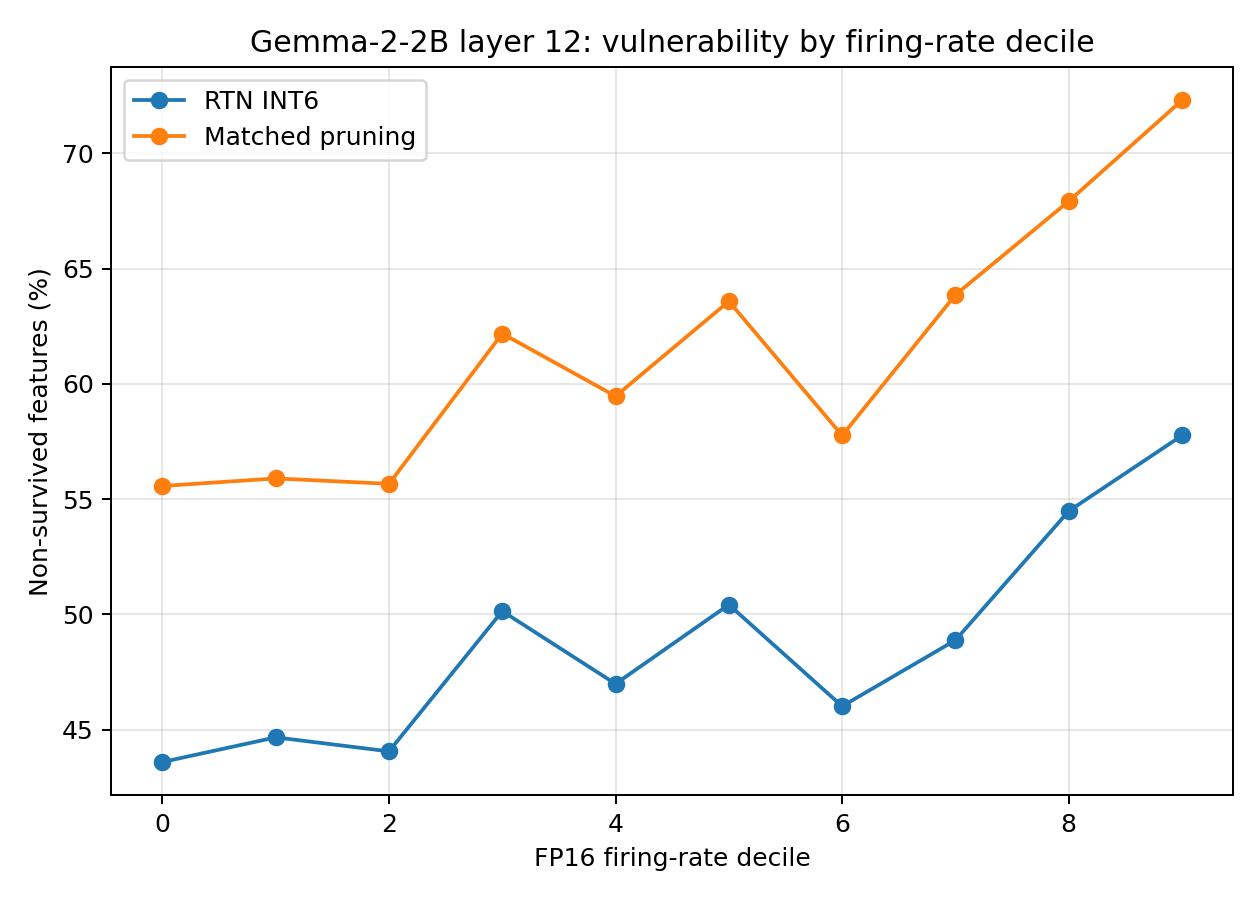}
    \caption{\gemmamodel{} non-survival rates by FP16 firing-rate decile for RTN INT6 and approximately matched pruning.}\label{fig:phase5b-gemma-rtn-vs-pruning-firing-rate-deciles}
\end{figure}

\clearpage
\section{Pruning calibration}\label{app:pruning-calibration}
\renewcommand{\thetable}{F.\arabic{table}}
\renewcommand{\theHtable}{F.\arabic{table}}
\setcounter{table}{0}

We calibrate magnitude pruning by searching over sparsity values \(p\in[0,0.8]\) to match the RTN INT6 perplexity regime. The search is performed with a fixed-step bisection procedure over six steps on a held-out token budget. For \pythiamodel{}, the selected pruning sparsity is \(p=0.175\), giving perplexity \(76.99\) and \(\Delta\mathrm{PPL}=+13.49\%\), close to RTN INT6 perplexity \(77.83\) and \(\Delta\mathrm{PPL}=+14.72\%\). For \gemmamodel{}, the selected pruning sparsity is \(p=0.1625\), giving perplexity \(486.63\) and \(\Delta\mathrm{PPL}=+6.13\%\), compared with RTN INT6 perplexity \(476.82\) and \(\Delta\mathrm{PPL}=+3.99\%\). Thus, the Pythia pruning baseline is closely matched, while the Gemma pruning baseline is approximately matched and slightly harsher.

\begin{table}[H]
    \centering
    \small
    \caption{\pythiamodel{} pruning calibration search for matching the RTN INT6 perplexity regime.}\label{tab:pythia-pruning-calibration-search}
    \begin{tabular}{cccc}
        \toprule
        Step & Sparsity & Perplexity & Delta (\%) \\
        \midrule
        0 & 0.4000 & 282.41 & 316.29 \\
        1 & 0.2000 & 82.77 & 22.01 \\
        2 & 0.1000 & 69.90 & 3.04 \\
        3 & 0.1500 & 74.25 & 9.45 \\
        4 & 0.1750 & 76.99 & 13.49 \\
        5 & 0.1875 & 78.75 & 16.08 \\
        \bottomrule
    \end{tabular}
\end{table}

\begin{table}[H]
    \centering
    \scriptsize
    \caption{\gemmamodel{} pruning calibration search for approximately matching the RTN INT6 perplexity regime.}\label{tab:gemma-pruning-calibration-search}
    \resizebox{\linewidth}{!}{%
    \begin{tabular}{ccccccccc}
        \toprule
        Step & Sparsity & Perplexity & Loss & Target PPL & Absolute error & Tensors & Params & Zeroed params \\
        \midrule
        0 & 0.4000 & 8,140.18 & 9.0046 & 497.99 & 7,642.19 & 182 & 2,024,275,968 & 811,793,027 \\
        1 & 0.2000 & 585.27 & 6.3721 & 497.99 & 87.28 & 182 & 2,024,275,968 & 405,794,737 \\
        2 & 0.1000 & 460.93 & 6.1333 & 497.99 & 37.06 & 182 & 2,024,275,968 & 202,895,355 \\
        3 & 0.1500 & 484.27 & 6.1827 & 497.99 & 13.71 & 182 & 2,024,275,968 & 304,557,827 \\
        4 & 0.1750 & 530.74 & 6.2743 & 497.99 & 32.76 & 182 & 2,024,275,968 & 355,143,150 \\
        5 & 0.1625 & 508.64 & 6.2317 & 497.99 & 10.65 & 182 & 2,024,275,968 & 329,809,692 \\
        \bottomrule
    \end{tabular}%
    }
\end{table}

\clearpage
\section{Fragility predictor details}\label{app:fragility-predictor-details}
\renewcommand{\thefigure}{G.\arabic{figure}}
\renewcommand{\thetable}{G.\arabic{table}}
\renewcommand{\theHfigure}{G.\arabic{figure}}
\renewcommand{\theHtable}{G.\arabic{table}}
\renewcommand{\figurename}{Appendix Figure}
\renewcommand{\tablename}{Appendix Table}
\setcounter{figure}{0}
\setcounter{table}{0}

Appendix G reports additional details for the fragility-prediction analysis: multivariate logistic-regression coefficients, cross-validated ROC curves, and a cross-model coefficient comparison. Because the predictors are correlated, the coefficients should be interpreted jointly rather than as isolated causal effects.

\begin{figure}[H]
    \centering
    \includegraphics[width=0.92\linewidth]{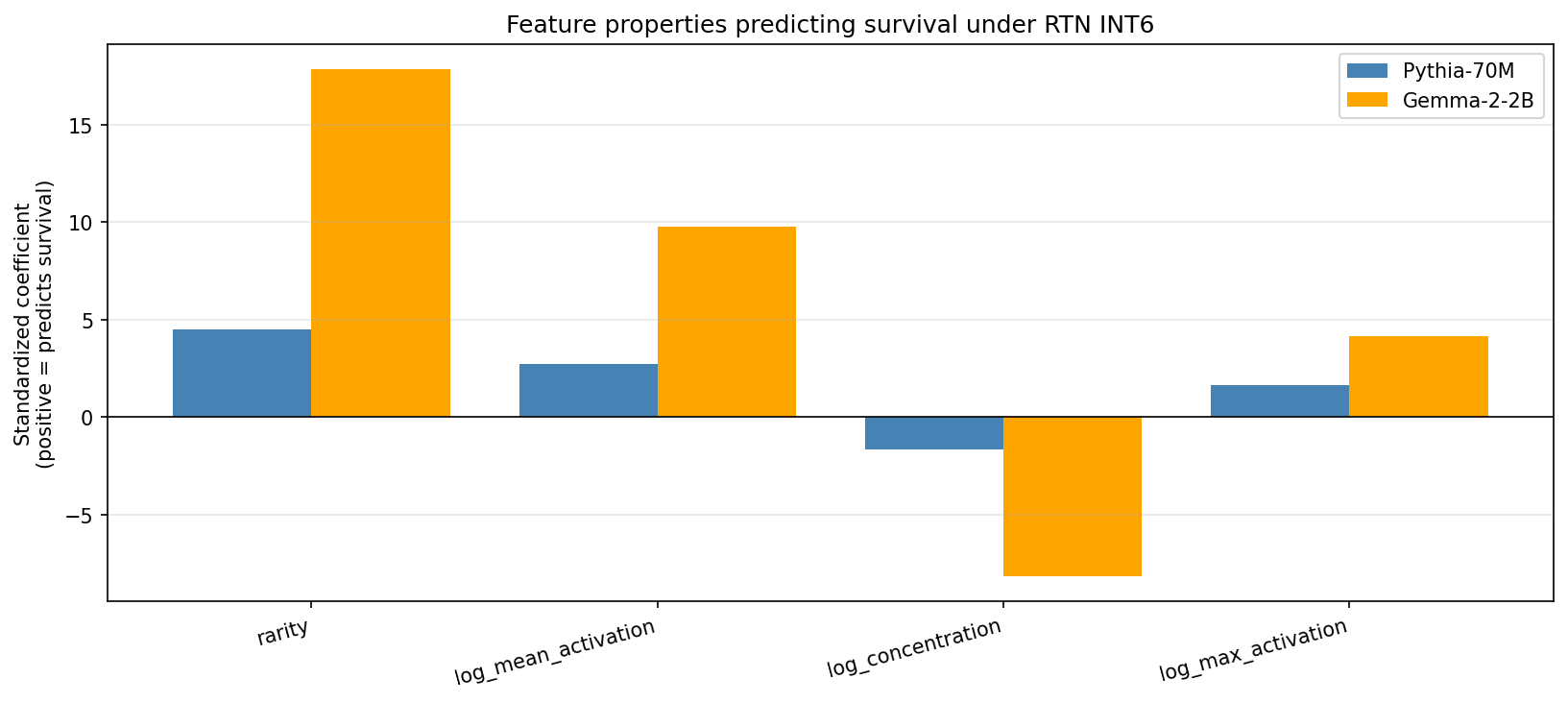}
    \caption{Standardized logistic-regression coefficients for predicting INT6 feature survival from FP16 feature statistics.}\label{fig:coefficient-comparison}
\end{figure}

\begin{figure}[H]
    \centering
    \includegraphics[width=0.92\linewidth]{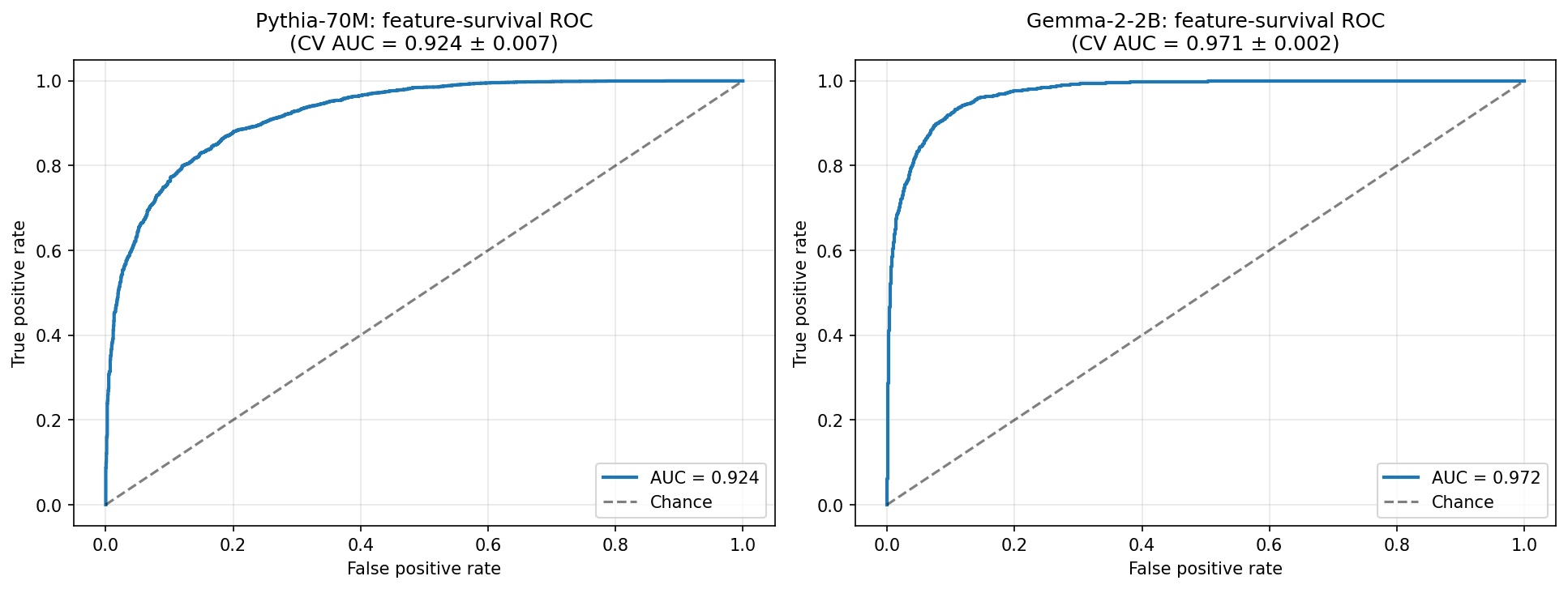}
    \caption{Cross-validated ROC curves for INT6 feature-survival prediction on \pythiamodel{} and \gemmamodel{}.}\label{fig:roc-curves}
\end{figure}

\begin{table}[H]
    \centering
    \small
    \caption{Cross-model comparison of standardized fragility-predictor coefficients.}\label{tab:coefficient-comparison}
    \begin{tabular}{lrrrr}
        \toprule
        Feature statistic & Pythia coef. & Gemma coef. & Sign agreement & Mean $|$coef.$|$ \\
        \midrule
        Rarity & 4.52 & 17.84 & Yes & 11.18 \\
        Log mean activation & 2.72 & 9.79 & Yes & 6.26 \\
        Log concentration & -1.67 & -8.16 & Yes & 4.92 \\
        Log peak activation & 1.66 & 4.14 & Yes & 2.90 \\
        \bottomrule
    \end{tabular}
\end{table}

\clearpage
\section{Condition-level verification summaries}\label{app:condition-level-verification-summaries}
\renewcommand{\thetable}{H.\arabic{table}}
\renewcommand{\theHtable}{H.\arabic{table}}
\renewcommand{\tablename}{Appendix Table}
\setcounter{table}{0}

Appendix H lists condition-level verification summaries for the main RTN INT6, pruning, and layer-sensitivity results. These files support the aggregate values reported in the main tables and figures.

\begin{table}[H]
    \centering
    \scriptsize
    \setlength{\tabcolsep}{3pt}
    \caption{\pythiamodel{} layer 4 summary statistics for the Phase 2A RTN quantization sweep.}\label{tab:phase2a-l4-summary}
    \textit{Feature-survival statistics}\\[0.25em]
    \begin{tabular}{lccccccccc}
        \toprule
        Condition & Bits & Layer & Tokens & Total feats. & Active feats. & Mean corr. & Median corr. & Survived (\%) & Damaged (\%) \\
        \midrule
        FP16 baseline & 16 & 4 & 199,680 & 32,768 & 5,686 & 1.0000 & 1.0000 & 100.00 & 0.00 \\
        per-channel INT8 & 8 & 4 & 199,680 & 32,768 & 5,686 & 0.9813 & 0.9920 & 98.33 & 0.00 \\
        per-channel INT7 & 7 & 4 & 199,680 & 32,768 & 5,686 & 0.9574 & 0.9804 & 86.32 & 0.00 \\
        per-channel INT6 & 6 & 4 & 199,680 & 32,768 & 5,686 & 0.8880 & 0.9395 & 62.40 & 1.00 \\
        per-channel INT5 & 5 & 4 & 199,680 & 32,768 & 5,686 & 0.7773 & 0.8497 & 37.95 & 12.57 \\
        \bottomrule
    \end{tabular}

    \smallskip
    \textit{Behavioral and degraded-feature statistics}\\[0.25em]
    \begin{tabular}{lccccc}
        \toprule
        Condition & Degraded (\%) & PPL & Loss & PPL delta (\%) \\
        \midrule
        FP16 baseline & 0.00 & 67.80 & 4.2166 & 0.00 \\
        per-channel INT8 & 1.67 & 68.49 & 4.2267 & 1.02 \\
        per-channel INT7 & 13.68 & 69.34 & 4.2390 & 2.27 \\
        per-channel INT6 & 36.60 & 77.78 & 4.3538 & 14.71 \\
        per-channel INT5 & 49.47 & 116.05 & 4.7540 & 71.16 \\
        \bottomrule
    \end{tabular}
\end{table}

\begin{table}[H]
    \centering
    \scriptsize
    \setlength{\tabcolsep}{3pt}
    \caption{\pythiamodel{} layer 2 summary statistics for the Phase 2A.5 layer-sensitivity check.}\label{tab:phase2a5-l2-summary}
    \textit{Feature-survival statistics}\\[0.25em]
    \begin{tabular}{lccccccccc}
        \toprule
        Condition & Bits & Layer & Tokens & Total feats. & Active feats. & Mean corr. & Median corr. & Survived (\%) & Damaged (\%) \\
        \midrule
        FP16 baseline & 16 & 2 & 199,680 & 32,768 & 6,139 & 1.0000 & 1.0000 & 100.00 & 0.00 \\
        per-channel INT8 & 8 & 2 & 199,680 & 32,768 & 6,139 & 0.9994 & 0.9996 & 100.00 & 0.00 \\
        per-channel INT7 & 7 & 2 & 199,680 & 32,768 & 6,139 & 0.9978 & 0.9985 & 100.00 & 0.00 \\
        per-channel INT6 & 6 & 2 & 199,680 & 32,768 & 6,139 & 0.9915 & 0.9942 & 99.95 & 0.00 \\
        per-channel INT5 & 5 & 2 & 199,680 & 32,768 & 6,139 & 0.9644 & 0.9753 & 94.67 & 0.07 \\
        \bottomrule
    \end{tabular}

    \smallskip
    \textit{Behavioral and degraded-feature statistics}\\[0.25em]
    \begin{tabular}{lccccc}
        \toprule
        Condition & Degraded (\%) & PPL & Loss & PPL delta (\%) \\
        \midrule
        FP16 baseline & 0.00 & 67.80 & 4.2166 & 0.00 \\
        per-channel INT8 & 0.00 & 68.49 & 4.2267 & 1.02 \\
        per-channel INT7 & 0.00 & 69.34 & 4.2390 & 2.27 \\
        per-channel INT6 & 0.05 & 77.78 & 4.3538 & 14.71 \\
        per-channel INT5 & 5.26 & 116.05 & 4.7540 & 71.16 \\
        \bottomrule
    \end{tabular}
\end{table}

\begin{table}[H]
    \centering
    \scriptsize
    \setlength{\tabcolsep}{3pt}
    \caption{\pythiamodel{} RTN INT6 per-condition summary used to verify the main INT6 results.}\label{tab:pythia-rtn-int6-verification-summary}
    \textit{Condition: RTN INT6}\\[0.25em]
    \begin{tabular}{cccccccc}
        \toprule
        Bits & Layer & Tokens & Active feats. & Mean corr. & Median corr. & Survived (\%) & Damaged (\%) \\
        \midrule
        6 & 4 & 199,680 & 5,688 & 0.8879 & 0.9395 & 62.39 & 1.02 \\
        \bottomrule
    \end{tabular}

    \smallskip
    \begin{tabular}{cccc}
        \toprule
        PPL & Loss & Method & Prune sparsity \\
        \midrule
        77.83 & 4.3545 & TL & \textemdash{} \\
        \bottomrule
    \end{tabular}
\end{table}

\begin{table}[H]
    \centering
    \scriptsize
    \setlength{\tabcolsep}{3pt}
    \caption{\pythiamodel{} matched-pruning per-condition summary used to verify the pruning baseline.}\label{tab:pythia-matched-pruning-verification-summary}
    \textit{Condition: Magnitude pruning matched INT6}\\[0.25em]
    \begin{tabular}{cccccccc}
        \toprule
        Bits & Layer & Tokens & Active feats. & Mean corr. & Median corr. & Survived (\%) & Damaged (\%) \\
        \midrule
        16 & 4 & 199,680 & 5,688 & 0.8905 & 0.9329 & 60.97 & 0.30 \\
        \bottomrule
    \end{tabular}

    \smallskip
    \begin{tabular}{cccc}
        \toprule
        PPL & Loss & Method & Prune sparsity \\
        \midrule
        76.99 & 4.3437 & TL & 0.1750 \\
        \bottomrule
    \end{tabular}
\end{table}

\begin{table}[H]
    \centering
    \scriptsize
    \setlength{\tabcolsep}{3pt}
    \caption{\gemmamodel{} RTN INT6 per-condition summary used to verify the main Gemma INT6 results.}\label{tab:gemma-rtn-int6-verification-summary}
    \textit{Condition: RTN INT6}\\[0.25em]
    \begin{tabular}{cccccccccc}
        \toprule
        Bits & Layer & Tokens & Total feats. & Active feats. & Mean corr. & Median corr. & Survived (\%) & Degraded (\%) & Damaged (\%) \\
        \midrule
        6 & 12 & 499,968 & 16,384 & 7,144 & 0.8911 & 0.9027 & 51.30 & 48.70 & 0.00 \\
        \bottomrule
    \end{tabular}

    \smallskip
    \begin{tabular}{cccc}
        \toprule
        PPL & Loss & PPL delta (\%) & Method \\
        \midrule
        476.82 & 6.1671 & 3.99 & RTN \\
        \bottomrule
    \end{tabular}
\end{table}

\begin{table}[H]
    \centering
    \scriptsize
    \setlength{\tabcolsep}{3pt}
    \caption{\gemmamodel{} approximately matched-pruning per-condition summary used to verify the Gemma pruning baseline.}\label{tab:gemma-matched-pruning-verification-summary}
    \textit{Condition: Magnitude pruning matched INT6}\\[0.25em]
    \begin{tabular}{ccccccccc}
        \toprule
        Layer & Tokens & Total feats. & Active feats. & Mean corr. & Median corr. & Survived (\%) & Degraded (\%) & Damaged (\%) \\
        \midrule
        12 & 499,968 & 16,384 & 7,144 & 0.8407 & 0.8551 & 38.58 & 61.18 & 0.24 \\
        \bottomrule
    \end{tabular}

    \smallskip
    \begin{tabular}{ccccccc}
        \toprule
        PPL & Loss & PPL delta (\%) & Method & Matched bits & Sparsity & Zeroed params \\
        \midrule
        486.63 & 6.1875 & 6.13 & pruning & 6 & 0.1625 & 329,809,692 \\
        \bottomrule
    \end{tabular}
\end{table}

\begin{table}[H]
    \centering
    \scriptsize
    \setlength{\tabcolsep}{2pt}
    \caption{Full combined cross-model RTN quantization and pruning summary.}\label{tab:combined-cross-model-rtn-pruning-summary}
    \resizebox{\linewidth}{!}{%
    \begin{tabular}{llrrrrrrrrrrrr}
        \toprule
        Model & Condition & Layer & Bits & Sparsity & Tokens & Active feats. & PPL & $\Delta$PPL (\%) & Mean corr. & Median corr. & Survived (\%) & Degraded (\%) & Damaged (\%) \\
        \midrule
        \pythiamodel{} & FP16 baseline & 4 & 16 & \textemdash{} & 199,680 & 5,688 & 67.840 & 0.000 & 1.000 & 1.000 & 100.000 & 0.000 & 0.000 \\
        \pythiamodel{} & RTN INT8 & 4 & 8 & \textemdash{} & 199,680 & 5,688 & 68.532 & 1.020 & 0.981 & 0.992 & 98.312 & 1.688 & 0.000 \\
        \pythiamodel{} & RTN INT7 & 4 & 7 & \textemdash{} & 199,680 & 5,688 & 69.372 & 2.258 & 0.957 & 0.980 & 86.322 & 13.678 & 0.000 \\
        \pythiamodel{} & RTN INT6 & 4 & 6 & \textemdash{} & 199,680 & 5,688 & 77.828 & 14.722 & 0.888 & 0.940 & 62.395 & 36.586 & 1.020 \\
        \pythiamodel{} & RTN INT5 & 4 & 5 & \textemdash{} & 199,680 & 5,688 & 116.148 & 71.208 & 0.777 & 0.850 & 37.957 & 49.455 & 12.588 \\
        \pythiamodel{} & RTN INT4 & 4 & 4 & \textemdash{} & 199,680 & 5,688 & 319.602 & 371.112 & 0.583 & 0.636 & 14.504 & 47.785 & 37.711 \\
        \pythiamodel{} & Magnitude pruning matched INT6 & 4 & 16 & 0.175 & 199,680 & 5,688 & 76.990 & 13.488 & 0.891 & 0.933 & 60.970 & 38.731 & 0.299 \\
        \midrule
        \gemmamodel{} & FP16 baseline & 12 & 16 & \textemdash{} & 499,968 & 7,144 & 458.535 & 0.000 & 1.000 & 1.000 & 100.000 & 0.000 & 0.000 \\
        \gemmamodel{} & RTN INT8 & 12 & 8 & \textemdash{} & 499,968 & 7,144 & 470.565 & 2.624 & 0.966 & 0.971 & 99.090 & 0.910 & 0.000 \\
        \gemmamodel{} & RTN INT7 & 12 & 7 & \textemdash{} & 499,968 & 7,144 & 432.636 & -5.648 & 0.943 & 0.950 & 81.257 & 18.743 & 0.000 \\
        \gemmamodel{} & RTN INT6 & 12 & 6 & \textemdash{} & 499,968 & 7,144 & 476.824 & 3.989 & 0.891 & 0.903 & 51.302 & 48.698 & 0.000 \\
        \gemmamodel{} & RTN INT5 & 12 & 5 & \textemdash{} & 499,968 & 7,144 & 539.768 & 17.716 & 0.764 & 0.778 & 27.716 & 66.363 & 5.921 \\
        \gemmamodel{} & RTN INT4 & 12 & 4 & \textemdash{} & 499,968 & 7,144 & 669.833 & 46.081 & 0.494 & 0.451 & 12.402 & 32.895 & 54.703 \\
        \gemmamodel{} & Magnitude pruning matched INT6 & 12 & \textemdash{} & 0.162 & 499,968 & 7,144 & 486.626 & 6.126 & 0.841 & 0.855 & 38.578 & 61.184 & 0.238 \\
        \bottomrule
    \end{tabular}%
    }
\end{table}

\clearpage
\section{Streaming estimator for per-feature correlations}\label{app:streaming-estimator}
\renewcommand{\theequation}{I.\arabic{equation}}
\renewcommand{\theHequation}{I.\arabic{equation}}
\setcounter{equation}{0}

Computing the feature-stability score in Eq.~\ref{eq:feature-correlation} naively requires storing the full activation matrices
$Z^{\mathrm{FP16}}, Z^{C} \in \mathbb{R}^{N \times d_{\mathrm{sae}}}$.
This is memory-intensive at the token budgets used in our experiments, especially for \gemmamodel{} where $N \approx 500{,}000$ and $d_{\mathrm{sae}}=16{,}384$.
We therefore compute the same Pearson correlations in a single streaming pass over batches of token positions.

For each feature $j$, define
\[
x_t = z_j(t;\theta_{\mathrm{FP16}}),
\qquad
 y_t = z_j(t;\theta_C).
\]
During streaming evaluation, we accumulate the following sufficient statistics:
\begin{equation}
\begin{aligned}
S_x &= \sum_t x_t,
& S_y &= \sum_t y_t,\\
S_{x^2} &= \sum_t x_t^2,
& S_{y^2} &= \sum_t y_t^2,\\
S_{xy} &= \sum_t x_t y_t.
\end{aligned}
\label{eq:streaming-sufficient-statistics}
\end{equation}
The Pearson correlation for feature $j$ is then recovered as
\begin{equation}
c_j =
\frac{S_{xy} - S_x S_y / N}
{\sqrt{\left(S_{x^2} - S_x^2/N\right)
\left(S_{y^2} - S_y^2/N\right)}}.
\label{eq:streaming-correlation}
\end{equation}
This expression is algebraically equivalent to Eq.~\ref{eq:feature-correlation}, but reduces memory from
$O(Nd_{\mathrm{sae}})$ to $O(d_{\mathrm{sae}})$.

We accumulate all sufficient statistics in float64 to reduce numerical error in the sum-of-squares computation.
In addition to the correlation statistics, we stream the FP16 firing count
\[
\sum_t \mathbf{1}\!\left[x_t>0\right]
\]
and the FP16 running maximum
\[
\max_t x_t
\]
used for the active-feature filter and feature-property analyses.
As a pipeline null check, comparing FP16 activations against themselves yields $c_j=1$ for all active features to six decimal places.

\clearpage
\section{Notation and Glossary}\label{app:notation-glossary}
\renewcommand{\thetable}{J.\arabic{table}}
\renewcommand{\theHtable}{J.\arabic{table}}
\renewcommand{\tablename}{Appendix Table}
\setcounter{table}{0}
\leftalignedcaptionstyle

This appendix summarizes the notation and threshold conventions used throughout the paper.

\begin{table}[H]
\centering
\caption{Notation and threshold conventions used in QDM feature-survival analysis.}\label{tab:notation-glossary}
\small
\begin{tabular}{>{\raggedright\arraybackslash}p{0.24\linewidth} >{\raggedright\arraybackslash}p{0.68\linewidth}}
\toprule
\textbf{Term / notation} & \textbf{Meaning} \\
\midrule
QDM & Our feature-level audit for measuring how SAE features change under model compression. QDM compares FP16 and compressed SAE activations on identical tokens using a fixed SAE. \\
FP16 model & The full-precision reference model. All feature-survival scores are measured relative to this model. \\
Compressed model & A model obtained by applying RTN quantization or magnitude pruning to the FP16 model weights. \\
RTN quantization & Round-to-nearest weight quantization. In our experiments, weights are rounded to a simulated $b$-bit grid for $b \in \{8,7,6,5,4\}$. \\
Matched pruning & Magnitude pruning calibrated to approximately match the RTN INT6 perplexity regime. It is used as a comparison baseline for quantization. \\
Read-out site & The residual-stream layer or hook where model activations are extracted and passed through the SAE. \\
Frozen SAE & The same pretrained SAE encoder is used for FP16 and compressed activations. The SAE is not retrained for compressed models. \\
$z_j(t;\theta)$ & Activation of SAE feature $j$ at token position $t$ for model parameters $\theta$. \\
$f_j$ & FP16 firing rate of feature $j$, defined as the fraction of evaluated token positions where $z_j(t;\theta_{\mathrm{FP16}})>0$. \\
Active feature & A feature with FP16 firing rate $f_j > 0.001$. Survival, degradation, and damage percentages are reported over active features unless stated otherwise. \\
$c_j$ & Feature-stability score for feature $j$, defined as the Pearson correlation between FP16 and compressed activations over identical token positions. \\
Survived feature & An active feature with $c_j > 0.9$, meaning its compressed activation pattern remains strongly aligned with its FP16 activation pattern. \\
Degraded feature & An active feature with $0.5 \leq c_j \leq 0.9$, meaning its compressed activation pattern remains partially but not strongly aligned with FP16. \\
Damaged feature & An active feature with $c_j < 0.5$, meaning its compressed activation pattern is weakly aligned with FP16 under the frozen SAE basis. \\
Feature fidelity & The degree to which SAE activation patterns are preserved after compression, measured primarily by $c_j$ and the survived/degraded/damaged taxonomy. \\
Behavioral metric & A task-level metric such as perplexity or downstream accuracy. Behavioral metrics are distinct from feature-level fidelity metrics. \\
$\Delta\mathrm{PPL}$ & Relative perplexity change compared with FP16, defined as $\mathrm{PPL}(C)/\mathrm{PPL}(\mathrm{FP16}) - 1$ for condition $C$. \\
Damage score & The per-feature quantity $d_j = 1 - c_j$, used when comparing how similarly RTN quantization and pruning affect individual features. \\
Jaccard overlap & The overlap between non-survived feature sets under two compression methods, defined as $|A \cap B|/|A \cup B|$. \\
\bottomrule
\end{tabular}
\end{table}

\clearpage
\section{Model, SAE, and Evaluation Configuration}\label{app:model-sae-evaluation-configuration}
\renewcommand{\thetable}{K.\arabic{table}}
\renewcommand{\theHtable}{K.\arabic{table}}
\renewcommand{\tablename}{Appendix Table}
\setcounter{table}{0}
\leftalignedcaptionstyle

This appendix summarizes the model, SAE, and evaluation settings used in the main QDM experiments.

\begin{table}[H]
\centering
\caption{Model, SAE, and token-budget configuration for the main QDM experiments. The SAE, token set, and read-out site are fixed within each model; only the model weights vary across compression conditions.}\label{tab:model-sae-configuration}
\small
\resizebox{\linewidth}{!}{%
\begin{tabular}{llllrr}
\toprule
Model & SAE source & Read-out site & SAE width & Token budget & Block length \\
\midrule
\pythiamodel{} & \texttt{pythia-70m-deduped-res-sm} & Layer 4 residual stream, \texttt{blocks.4.hook\_resid\_post} & \textemdash{} & 200k & 512 \\
\gemmamodel{} & Gemma Scope canonical residual-stream SAE & Layer 12 residual stream & 16,384 & 500k & 256 \\
\bottomrule
\end{tabular}%
}
\end{table}

\begin{table}[H]
\centering
\caption{Shared experimental conventions for QDM feature-survival evaluation.}\label{tab:shared-experimental-conventions}
\small
\begin{tabular}{>{\raggedright\arraybackslash}p{0.28\linewidth} >{\raggedright\arraybackslash}p{0.64\linewidth}}
\toprule
\textbf{Setting} & \textbf{Value / description} \\
\midrule
Dataset & WikiText-2-raw. Non-empty lines are concatenated and tokenized before evaluation. \\
Compression methods & Round-to-nearest quantization and magnitude pruning. \\
RTN bit-widths & INT8, INT7, INT6, INT5, and INT4. \\
Quantized modules & Attention projections and MLP projections; for \gemmamodel{}, the gated MLP gate projection is also quantized. Layer-norm and embedding parameters are left in full precision. \\
Quantization granularity & Per-output-channel RTN. Quantized weights are dequantized back into floating-point tensors for simulated low-bit evaluation. \\
Pruning baseline & Magnitude pruning calibrated to approximately match the RTN INT6 perplexity regime. \\
Active-feature threshold & A feature is active if its FP16 firing rate satisfies $f_j > 0.001$. \\
Survival threshold & A feature survives if its FP16-vs-compressed activation correlation satisfies $c_j > 0.9$. \\
Damage threshold & A feature is damaged if its FP16-vs-compressed activation correlation satisfies $c_j < 0.5$. \\
Default survival taxonomy & Survived: $c_j > 0.9$; degraded: $0.5 \leq c_j \leq 0.9$; damaged: $c_j < 0.5$. \\
Feature-stability metric & Pearson correlation $c_j$ between FP16 and compressed SAE activations over identical token positions. \\
Behavioral metric & Token-level perplexity, reported as relative change $\Delta\mathrm{PPL}$ from FP16. \\
Sliding-window check & \gemmamodel{} behavioral robustness check with window size $W=2048$ and stride $S=512$. \\
Fragility predictor & L2-regularized logistic regression predicting INT6 survival from FP16 feature statistics. \\
Stability checks & Token-budget sensitivity, random-subset stability, FP16-vs-FP16 null, threshold sensitivity, and second-layer sensitivity. \\
\bottomrule
\end{tabular}
\end{table}

\clearpage
\section{Quantized Module Implementation Details}\label{app:quantized-module-implementation-details}
\renewcommand{\thetable}{L.\arabic{table}}
\renewcommand{\theHtable}{L.\arabic{table}}
\renewcommand{\tablename}{Appendix Table}
\setcounter{table}{0}
\leftalignedcaptionstyle

This appendix specifies which weight tensors are modified by the compression operators. For both RTN quantization and magnitude pruning, we apply compression only to transformer block linear weights. Embeddings, unembeddings, layer-normalization parameters, biases, and SAE parameters are left unchanged.

For each transformer block, the targeted attention weights are the query, key, value, and output projections. The targeted MLP weights are the input/up projection and output/down projection. For \gemmamodel{}, which uses a gated MLP, we additionally compress the gate projection. Thus, the compressed module families are:

\begin{table}[H]
\centering
\caption{Transformer block module families included in, and excluded from, compression.}\label{tab:quantized-module-families}
\small
\begin{tabular}{>{\raggedright\arraybackslash}p{0.28\linewidth} >{\raggedright\arraybackslash}p{0.64\linewidth}}
\toprule
\textbf{Component} & \textbf{Targeted weights} \\
\midrule
Attention & Query projection, key projection, value projection, output projection. \\
MLP & Input/up projection, output/down projection. \\
Gated MLP, Gemma only & Gate projection. \\
Excluded & Embeddings, unembeddings, layer norms, biases, SAE weights. \\
\bottomrule
\end{tabular}
\end{table}

In TransformerLens-style notation, the Pythia targeted modules correspond to the attention projection weights and MLP projection weights in each block, including names such as \texttt{W\_Q}, \texttt{W\_K}, \texttt{W\_V}, \texttt{W\_O}, \texttt{W\_in}, and \texttt{W\_out}. In HuggingFace/Gemma-style notation, the targeted modules correspond to names such as \texttt{q\_proj}, \texttt{k\_proj}, \texttt{v\_proj}, \texttt{o\_proj}, \texttt{up\_proj}, \texttt{down\_proj}, and \texttt{gate\_proj}.

All targeted tensors are compressed with the same per-output-channel procedure described in Section~\ref{subsec:compression-operators}. For RTN quantization, weights are rounded to the target integer grid and then dequantized back into floating-point tensors, so the experiment simulates low-bit weights without changing the model architecture. For pruning, the smallest-magnitude weights within each targeted tensor are set to zero according to the calibrated sparsity. The SAE encoder is never quantized, pruned, or retrained.

The output-channel convention follows the tensor layout used by the loaded model implementation. For each targeted matrix, the scale is computed over all dimensions except the output-channel dimension, so each output channel receives its own RTN scale. This convention is applied consistently across Pythia and Gemma experiments; where module layouts differ between TransformerLens and HuggingFace implementations, the corresponding output-channel axis is selected before applying the same per-output-channel rule.

\section{Reproducibility and Hardware}\label{app:reproducibility-hardware}

We provide an anonymized code and artifact repository at \url{https://anonymous.4open.science/r/sae-feature-survival-quantization-0141/}. The repository includes the experiment scripts, configuration files, lightweight utility code, generated summary outputs, and figures used to reproduce the main analyses. The scripts cover the RTN bit-width sweeps, streaming feature-correlation pipeline, Gemma sliding-window perplexity check, stability ablations, fragility-predictor analysis, and table/figure generation. Large model weights, datasets, and third-party SAE checkpoints are not redistributed; they should be downloaded from their original sources subject to their respective licenses. The repository is anonymized for review and does not include author-identifying metadata.

Experiments were run on CUDA-enabled A100 GPU hardware. Approximately 24 hours of total computation time including debugging. The Pythia-70M experiments are lightweight and can be reproduced on a single consumer GPU with sufficient memory, while the Gemma-2-2B streaming and sliding-window evaluations require a larger GPU or careful batching/offloading. To reduce memory usage, our implementation computes feature correlations from streaming sufficient statistics rather than storing the full token-by-feature activation matrix. The main experiments use 200k tokens for Pythia-70M and 500k tokens for Gemma-2-2B, with smaller-token-budget runs used for stability checks.

\end{document}